\documentclass[preprint, times]{elsarticle}
\usepackage[margin=3cm]{geometry}

\usepackage{amsmath,amsfonts,bm, amssymb}

\usepackage{microtype}
\usepackage{graphicx}
\usepackage{subcaption}
\usepackage{tikz}
\usetikzlibrary{positioning}
\usetikzlibrary{arrows,shapes,snakes,automata,backgrounds,fit,petri,calc}
\usepackage{booktabs} 
\usepackage{hyperref}
\usepackage{url}
\usepackage{xcolor}
\usepackage{wrapfig}
\usepackage{cleveref}
\usepackage[acronym,nowarn]{glossaries}
\usepackage[ruled,lined]{algorithm2e}

\newacronym{VAE}{VAE}{variational autoencoder}
\newacronym{PIDVAE}{$\Phi$-DVAE}{physics-informed dynamical variational autoencoder}
\newacronym{SSM}{SSM}{state-space model}
\newacronym{ELBO}{ELBO}{evidence lower bound}
\newacronym{VLB}{VLB}{variational lower bound}
\newacronym{VB}{VB}{variational Bayes}
\newacronym{MC}{MC}{Monte Carlo}

\newacronym{NN}{NN}{neural network}
\newacronym{PINN}{PINN}{physics-informed neural network}
\newacronym{GP}{GP}{Gaussian process}
\newacronym{DVAE}{DVAE}{dynamical variational autoencoder}
\newacronym{KVAE}{KVAE}{Kalman variational autoencoder}
\newacronym{GPVAE}{GPVAE}{Gaussian process variational autoencoder}
\newacronym{VRNN}{VRNN}{variational recurrent neural networks}

\newacronym{ExKF}{ExKF}{extended Kalman filter}
\newacronym{DA}{DA}{data assimilation}
\newacronym{UDA}{UDA}{unstructured data assimilation}
\newacronym{SDE}{SDE}{stochastic differential equation}
\newacronym{ODE}{ODE}{ordinary differential equation}
\newacronym{PDE}{PDE}{partial differential equation}

\setacronymstyle{long-sc-short}
\newacronym{KdV}{KdV}{Korteweg-de Vries}
\newacronym{MSE}{MSE}{mean-squared-error}

\newacronym{FEM}{FEM}{finite element method}
\newacronym{statFEM}{statFEM}{statistical finite element method}

\usepackage{mathtools}
\DeclareMathOperator{\KLOP}{KL}
\DeclarePairedDelimiterX{\DistanceRelation}[2]{(}{)}{#1\;\delimsize\|\;#2}
\newcommand{\kl}{\KLOP\DistanceRelation}

\def\vb{{\mathbf{b}}}
\def\ve{{\mathbf{e}}}
\def\vh{{\mathbf{h}}}
\def\vm{{\mathbf{m}}}
\def\vr{{\mathbf{r}}}
\def\vs{{\mathbf{s}}}
\def\vu{{\mathbf{u}}}
\def\vw{{\mathbf{w}}}
\def\vx{{\mathbf{x}}}
\def\vy{{\mathbf{y}}}

\def\mA{{\mathbf{A}}}
\def\mC{{\mathbf{C}}}

\def\mG{{\mathbf{G}}}
\def\mH{{\mathbf{H}}}
\def\mI{{\mathbf{I}}}
\def\mJ{{\mathbf{J}}}
\def\mL{{\mathbf{L}}}
\def\mM{{\mathbf{M}}}
\def\mQ{{\mathbf{Q}}}
\def\mR{{\mathbf{R}}}
\def\mW{{\mathbf{W}}}
\def\mLambda{{\mathbf{\Lambda}}}

\def\md{{\mathrm d}}

\def\cA{{\mathcal A}}
\def\cF{{\mathcal F}}
\def\cM{{\mathcal M}}
\def\NPDF{{\mathcal N}}

\def\bR{{\mathbb R}}

\DeclareMathOperator{\diag}{diag}
\DeclareMathOperator{\spans}{span}

\SetKwInput{KwInput}{Input}                
\SetKwInput{KwOutput}{Output}              
\SetKwComment{Comment}{$\triangleright$\ }{}

\begin{document}

\begin{frontmatter}

\title{$\Phi$-DVAE: Physics-Informed Dynamical Variational Autoencoders for Unstructured Data Assimilation}

\author[cambridge]{Alex Glyn-Davies\corref{cor1}}
\cortext[cor1]{Corresponding author. Email: \texttt{ag933@cam.ac.uk}}

\author[cambridge]{Connor Duffin}

\author[imperial]{O. Deniz Akyildiz}

\author[cambridge,turing]{Mark Girolami}

\affiliation[cambridge]{
            organization={Department of Engineering, University of Cambridge},
            addressline={Trumpington St},
            city={Cambridge},
            postcode={CB2 1PZ},
            country={United Kingdom}}

\affiliation[imperial]{
            organization={Department of Mathematics, Imperial College London},
            addressline={Exhibition Rd},
            city={London},
            postcode={SW7 2AZ},
            country={United Kingdom}}

\affiliation[turing]{
            organization={The Alan Turing Institute},
            city={London},
            addressline={British Library, 96 Euston Rd},
            postcode={NW1 2DB},
            country={United Kingdom}}

\begin{abstract}
Incorporating unstructured data into physical models is a challenging problem that is emerging in data assimilation. Traditional approaches focus on well-defined observation operators whose functional forms are typically assumed to be known. This prevents these methods from achieving a consistent model-data synthesis in configurations where the mapping from data-space to model-space is unknown. To address these shortcomings, in this paper we develop a physics-informed dynamical variational autoencoder ($\Phi$-DVAE) to embed diverse data streams into time-evolving physical systems described by differential equations. Our approach combines a standard, possibly nonlinear, filter for the latent state-space model and a VAE, to assimilate the unstructured data into the latent dynamical system. Unstructured data, in our example systems, comes in the form of video data and velocity field measurements, however the methodology is suitably generic to allow for arbitrary unknown observation operators. A variational Bayesian framework is used for the joint estimation of the encoding, latent states, and unknown system parameters. To demonstrate the method, we provide case studies with the Lorenz-63 ordinary differential equation, and the advection and Korteweg-de Vries partial differential equations. Our results, with synthetic data, show that $\Phi$-DVAE provides a data efficient dynamics encoding methodology which is competitive with standard approaches. Unknown parameters are recovered with uncertainty quantification, and unseen data are accurately predicted.
\end{abstract}

\end{frontmatter}


\section{Introduction}
\label{sec:intro}
Physical models, as represented by differential equations, are ubiquitous throughout engineering and the physical sciences. These equations synthesise scientific knowledge into mathematical form. However, as a description of reality they are imperfect~\citep{judd2004indistinguishable}, leading to the well-known problem of model misspecification~\citep{box1979robustness}. At least since~\citet{kalman1960new} physical modellers have been trying to reconcile their inherently misspecified models with observations \citep{anderson1979optimal}. Such approaches are usually either solving the inverse problem of attempting to recover model parameters from data, and/or, the \gls*{DA} problem of conducting state inference based on a time-evolving process.

For the inverse problem, Bayesian methods are common~\citep{tarantola2005inverse,stuart2010inverse}. In this, prior belief in model parameters $\mLambda$ is updated with data $\vy$ to give a posterior distribution, $p(\mLambda | \vy)$. This describes uncertainty with parameters given the data and modelling assumptions. \gls*{DA} can also proceed from a Bayesian viewpoint, where inference is cast as a nonlinear \gls*{SSM}~\citep{law2015data, reich2015probabilistic}. The \gls*{SSM} is typically the combination of a time-discretised differential equation and an observation process: uncertainty enters the model through extrusive, additive errors. For a latent state variable $\vu_n$ representing some discretised system at time $n$, with observations $\vy_n$, the object of interest is the filtering distribution $p(\vu_n | \vy_{1:n})$, where $\vy_{1:n} := \{\vy_k\}_{k = 1}^n$. The joint filtering and estimation problem, which estimates $p(\vu_n, \mLambda | \vy_{1:n})$ has received attention in the literature (see, e.g., \citet{kantas2015particle} and references therein). This has been well studied in, e.g., electrical engineering \citep{storvik2002particle}, geophysics \citep{bocquet2013joint}, neuroscience \citep{ditlevsen2014estimation}, chemical engineering \citep{kravaris2013advances}, biochemistry \citep{dochain2003state}, and hydrology \citep{moradkhani2005dual}, to name a few.

Typically in data assimilation tasks, while parameters of an observation model may be unknown, the observation model itself is assumed known~\citep{kantas2015particle}. This assumption breaks down in settings where data arrives in various modalities, such as videos, images, or audio, hindering the ability to perform inference. However, in such cases often the underlying variation in the data stream is due to a latent physical process, which is typically at least partially known. We refer to this setting, where the observation operator of the physical process is unknown, as the problem of \gls*{UDA}. In many physics-based applications, simplified or partially known physics models are available, e.g. weather prediction, soil mechanics, fluid mechanics, but require data to be assimilated into the model to adjust for misspecification, and improve predictive capabilities. Our framework is developed with these applications in mind, where we take a physics-informed machine learning approach to allow for the assimilation of \textit{unstructured} data into underlying physical models.

In this work, we develop a \gls*{VB}~\citep{blei2017variational} methodology which jointly solves the inverse and filtering problems when this observation operator is unknown. We model this unknown mapping with a \gls*{VAE}~\citep{kingma2014autoencoding}, which embeds the assumed time-dependent observations $\vy_{1:N}$ into encodings $\vx_{1:N}$ in a latent space. On this latent space, we stipulate that the latent encodings are observations emitted from a known dynamical system, given by a stochastic differential equation with possibly unknown coefficients. Following \cite{fraccaro2017disentangled}, we refer to these latent encodings as ``pseudo-observations''. To be clear, these pseudo-observations are not directly observed, but are treated as observations of the latent \gls*{SSM}. The differential equation is also assumed to have stochastic forcing, which accounts for possible model misspecification. The stipulated system gives a structured prior $p(\vx_{1:N} | \mLambda)$, which acts as a physics-informed regulariser whilst also enabling inference over the unknown model parameters $\mLambda$. This prior is approximated using classical nonlinear filtering algorithms.  Our framework is fully probabilistic: inference proceeds from a derived \gls*{ELBO}, enabling joint estimation of unknown network parameters and unknown dynamical coefficients via \gls*{VB}. To set the scene for this work, we now review the relevant literature.

\section{Related Work}

As introduced above, \glspl{VAE} \citep{kingma2014autoencoding} are a popular high-dimensional encoder. A \gls*{VAE} defines a generative model that learns low-dimensional representations, $\vx$, of high-dimensional data, $\vy$, using \gls*{VB} in an unsupervised fashion. To perform efficient inference, a variational approximation $q_{\phi}(\vx|\vy)$ is made to the intractable posterior $p(\vx|\vy)$, via a \gls*{NN}. Variational parameters $\phi$ are estimated via the \gls*{ELBO}. Recent works have extended the \gls*{VAE} to high-dimensional time-series data $\vy_{1:N}$, with the aim of jointly learning latent representations $\vx_{1:N}$, and a dynamical system that evolves them~\citep{girin2021dynamical}. Termed \glspl*{DVAE}, they enforce the dynamics with a structured prior $p(\vx_{1:N})$ on the latent space.

The \gls*{KVAE} of~\cite{fraccaro2017disentangled} is a \gls*{DVAE} methodology, which encodes $\vy_{1:N}$ into latent variables $\vx_{1:N}$ that are assumed to be observations of a linear Gaussian state-space model driven by latent dynamic states $\vu_{1:N}$. Assumed linear dynamics are jointly learnt with the encoder and decoder, via Kalman filtering/smoothing. Another approach is the \gls*{GP} VAE~\citep{pearce2020gaussian,jazbec2021scalable,fortuin2020gp}, which models $\vx_{1:N}$ as a temporally correlated \gls*{GP}. The Markovian variant of~\cite{zhu2022markovian} allows for a similar Kalman procedure as in the \gls*{KVAE}, except, in this instance, the dynamics are known and are given by an \gls*{SDE} approximation to the \gls*{GP}~\citep{hartikainen2010kalman}. \citet{yildiz2019ode2vae} also propose the so-called ODE$^2$VAE, which encodes the data to an initial condition which is integrated through time using a Bayesian neural \gls*{ODE}~\citep{chen2018neural}. This trajectory, only, is used to generate the reconstructions via the decoder network. 

A related class of methods are deep \gls*{SSM}s~\citep{bayer2014learning,krishnan2015deep,karl2017deep}. These works assume that the parametric form of the SSM is unknown, and replace the transition and emission distributions with \gls*{NN} models, which are trained based on an ELBO. They harness the representational power of deep \glspl*{NN} to directly model transitions between high-dimensional states. More emphasis is placed on generative modelling and prediction than representation learning, or system identification. We also note the related works which use VAE-type architectures for video prediction tasks~\citep{wu2021greedy,franceschi2020stochastic,babaeizadeh2022stochastic},
and the variational recurrent approach of~\citet{chung2015recurrent}, which is predominantly demonstrated on speech modelling.

Methods to blend physics and autoencoders have also been developed. A popular approach uses SINDy~\citep{brunton2016discovering} for discovery of low-dimensional latent dynamical systems using autoencoders~\citep{champion2019data}. A predictive framework is given in~\citet{lopez2021variational}, which aims to learn nonlinear dynamics by jointly optimising an \gls*{ELBO}. Following our notation, this learns a function which maps $\vu_n \mapsto \vu_{n + k}$, for some $k$, via a \gls*{VAE}. \citet{lusch2018deep} use a physics-informed autoencoder to linearise nonlinear dynamical systems via a Koopman approach; inference is regularised through incorporating the Koopman structure in the loss function. \citet{otto2019linearly} present a similar method, and an extension of these approaches to PDE systems is given in~\citet{gin2021deep}. \citet{morton2018deep} use the linear regression methods of \citet{takeishi2017learning} within a standard autoencoder to similarly compute the Koopman observables. \citet{erichson2019physicsinformed} derive an autoencoder which incorporates a linear Markovian prediction operator, similar to a Koopman operator, which uses physics-informed regulariser to promote Lyapunov stability. \citet{hernandez2018variational} studies VAE methods to encode high-dimensional dynamical systems. Finally, we note the related work which studies the estimation of dynamical parameters within so-called ``gray-box'' systems, blending \gls*{NN} methods with known physical laws~\citep{lu2020extracting,yin2021augmenting,long2018pdeneta,bezenac2019deep}. Methods for solving Bayesian inverse problems using \glspl*{VAE} have been developed for stochastic differential equations \cite{SHIN2023112183}, and notably the UQ-VAE \cite{goh2022solving} for when the forward model is known. The UQ-VAE uses observation/latent pairs for training, and the forward model used as the decoder. These methods require the observation operator to be known.

Going beyond such ``gray-box'' methods, deep learning methods have been increasingly applied to directly approximating \glspl*{PDE}. Termed the \gls*{PINN}~\cite{RAISSI2019686}, physical laws corresponding to the \gls*{PDE} are imposed on the loss function of the \gls*{NN}; solutions that do not obey such laws are thus penalised. The framework is flexible, and can be used to solve high-dimensional forward problems \cite{Wandel20228529, ZHU201956}, or to solve inverse problems where noisy data are observed \cite{RAISSI2019686}. A Bayesian variant of the standard \gls*{PINN} has been developed in \cite{YANG2021109913} to provide uncertainty quantification in the forward and inverse problem setting, using Bayesian neural networks.

These \gls*{PINN} models fall under the category of ``physics-informed'' machine learning. Recent reviews of physics-informed machine learning \cite{meng2022physics, karniadakis2021physics} use this term to refer to integration of prior physics knowledge/mathematical physics models with data-driven machine learning methods. More specifically, \glspl*{PINN} and other methodologies that impose \gls*{PDE} constraints to learn surrogate models \textit{without} data are termed ``physics-constrained" learning methods \cite{ZHU201956, geneva2020modeling}, allowing surrogates to be constructed without the need for solving the governing equations. An emerging term in the dynamical systems literature is ``physics-enhanced" machine learning \cite{cicirello2024physics, lathourakis2024physics}. Coined as an umbrella term, this encompasses three types of scientific machine learning: physics-guided, physics-encoded, and physics-informed. Physics-guided having the highest level of physics bias, physics-informed being more data-driven, and physics-encoded providing a combination of both.

Another recent advancement in solving both forward and inverse problems using deep learning is the neural operator \cite{kovachki2023neural}. Neural operators are parameterised maps between function spaces, and have been used to learn \gls*{PDE} solution operators. They are therefore discretisation-invariant models, making them robust to mesh-refinement, and able to solve super-resolution problems \cite{li2020fourier}. 

These model can be used in place of slower, traditional PDE solvers as efficient surrogates for the forward model, and perform comparably when used for downstream tasks such as Bayesian inversion \cite{li2021physics}. Related variational approaches have also been developed for time-varying physical systems by parameterising time as another input to deep NNs that are used to parameterise the variational family \cite{vadeboncoeur2022deep, vadeboncoeur2023random}.

\subsection*{Our contribution} 

In this paper we propose a \gls*{PIDVAE}: a \gls*{DVAE} approach which imposes the additional structure of known physics on the latent space. The term physics-informed is used in this paper to describe this integration of the data-driven autoencoder model with the known mathematical physics model. We assume that there is a low-dimensional dynamical system generating the high-dimensional observed time-series. A \gls*{NN} is used to learn the unknown embedding to this lower dimensional space. On the lower-dimensional space, the embedded data are pseudo-observations of a latent dynamical system, which is, in general, derived from a numerical discretisation of a nonlinear PDE. However, the methodology is suitably generic, allowing for ODE latent systems. Inference on this latent system is done with efficient nonlinear stochastic filtering methods, enabling the use of mature \gls*{DA} algorithms within our framework. A probabilistically coherent \gls*{VB} construction allows for joint learning of both the embedding and unknown dynamical parameters.

Instead of learnt dynamics with the KVAE~\citep{fraccaro2017disentangled}, \gls*{PIDVAE} assumes a misspecified nonlinear differential equation is driving the variation in the latent space, with possibly unknown parameters. Specifying the dynamics gives a known latent transition density, which is parameterised by physics parameters instead of neural networks as in, e.g., deep \glspl*{SSM}~\citep{bayer2014learning,krishnan2015deep,karl2017deep}. Whilst we share commonality with latent differential equations, the \gls*{PIDVAE} differs with the ODE$^2$VAE~\citep{yildiz2019ode2vae} as we perform inference with this ODE/PDE, instead of learning it and leveraging it to deterministically evolve the latents. Specifying transitions via a differential equation differs the methodology to incorporating generic physics principles in the latent space~\citep[such as][]{lusch2018deep,otto2019linearly,gin2021deep,erichson2019physicsinformed}, or generic temporal structure~\citep[such as][]{pearce2020gaussian,jazbec2021scalable,fortuin2020gp,zhu2022markovian}.

Our work supplements the physics-informed machine learning literature through addressing the crucial gap of embedding a physics-driven description of latent state evolution within \glspl*{VAE}. We also provide, to the best of our knowledge, the first contribution within this field which addresses the problem of inversion using physics-informed machine learning \citep[seen in][]{SHIN2023112183, goh2022solving, RAISSI2019686, YANG2021109913, li2020fourier, li2021physics} under an unknown observation operator. Results, with synthetic data, show that the method learns sensible encodings, in limited-data settings, and  accurately estimates physical model parameters, with uncertainty quantification. We further show that the method performs comparably to current approaches, and, when extrapolating beyond training data, accurately predicts future latent states.

This paper is structured as follows. The probabilistic model is described in Section~\ref{sec:model}, which shows how the known physics are embedded into the latent space of a \gls*{VAE} (Section~\ref{subsec:dynamic-model}). The pseudo-observation model for the physical system is defined in Section~\ref{subsec:likelihood} and the pseudo-observations $\vx_{1:N}$ are related to the unstructured data $\vy_{1:N}$ in Section~\ref{subsec:decoder}. In Section~\ref{subsec:variational-inference} we derive the variational inference in the setting of the probabilistic model. The learning objective is given by the log-marginal likelihood of the observed data $\vy_{1:N}$ as defined by the probabilistic model, and we derive a tractable lower-bound approximation to this. In \Cref{subsec:joint-learning,subsec:filtering-inference} we define the inverse problem we wish to solve, and outline the algorithm for performing the inference. Experiments and results for three different examples are given in Section~\ref{sec:experiments}. The paper is concluded in Section~\ref{sec:conclusion}.

\section{The Probabilistic Model}
\label{sec:model}

In this section we define our probabilistic model; our presentation roughly
follows the structure of the generative model. We first give an overview of the
dependencies between variables, as described by conditional probabilities. We
then cover the latent differential equation model used to describe the
underlying physics. The pseudo-observation model is defined, followed by
the decoder and the encoder. We assume a
general \gls*{SSM}:
\begin{align*}
  \mLambda &\sim p(\mLambda), \\
  \vu_n | \vu_{n-1}, \mLambda &\sim p(\vu_n | \vu_{n-1}, \mLambda), \\
  \vx_n | \vu_{n} &\sim p(\vx_n | \vu_n), \\
  \vy_n | \vx_n &\sim p_\theta(\vy_n | \vx_n),
\end{align*}
where $\mLambda$ describes the parameters of the Markov process $\{\vu_n\}_{n=0}^N$. The \textit{dynamic model} $p(\vu_n | \vu_{n-1}, \mLambda)$ defines the evolution of this process in time. The pseudo-observation likelihood is denoted by $p(\vx_n | \vu_n)$. 
 Parameters $\theta$ are the \gls*{NN} parameters for the \textit{decoder} $p_\theta(\vy_n | \vx_n)$.  The conditional independence structure imposed by the model gives
\begin{align}
  \label{eq:factor-joint}
  p(\vy_{1:N}, \vx_{1:N}, \vu_{1:N}, \mLambda)
  &= p_{\theta}(\vy_{1:N} | \vx_{1:N}) p(\vx_{1:N} | \vu_{1:N}) p(\vu_{1:N} | \mLambda) p(\mLambda).
\end{align}
Intuitively, $\{\vy_n\}_{n = 1}^N$ is the sequence of high-dimensional observations, $\{\vx_n\}_{n=1}^N$ is its embedding/pseudo-data, and $\{\vu_n\}_{n = 0}^N$ is the \textit{latent physics process}. For each $n$, we assume that $\vy_n \in \bR^{n_y}$, $\vx_n \in \bR^{n_x}$, $\vu_n \in \bR^{n_u}$, and $\mLambda \in \bR^{n_\lambda}$. In what follows, we describe the components of our probabilistic model in detail.

\begin{figure}[t]
  \centering 
  \includegraphics[width=\textwidth]{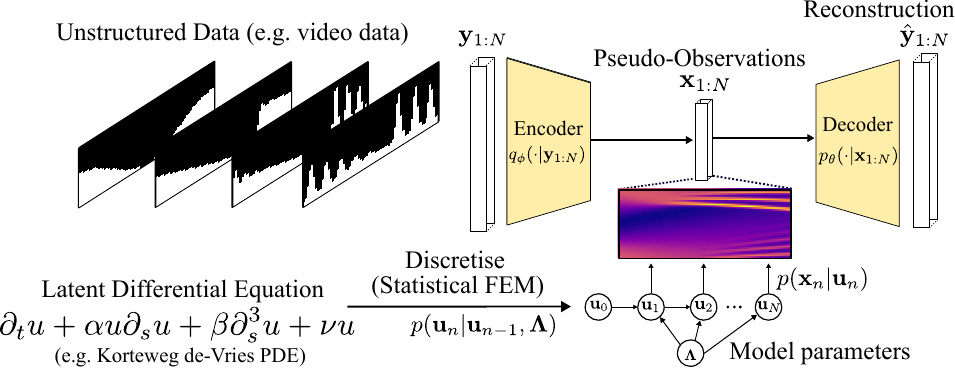}
  \caption{
  An illustration of the $\Phi$-DVAE model. On the left, the video frames are seen, denoted $\vy_{1:N}$. These are converted into physically interpretable low-dimensional encodings $\vx_{1:N}$ using an encoder. The learning is informed by the physics-driven state-space model, which treats $\vx_{1:N}$ as \textit{pseudo-observations} (bottom right). These pseudo-observations are used to infer the latent states $\vu_{1:N}$.}
  \label{fig:network}
 \end{figure}

\subsection{Dynamic Model}
\label{subsec:dynamic-model}
The first component of the generative model is the latent dynamical system
$p(\vu_n | \vu_{n-1}, \mLambda)$. We model this latent physics process
$\{\vu_n\}_{n = 0}^N$ as a discretised stochastic \gls*{PDE}, however \gls*{ODE} latent
physics is admissible within our framework (see~Section~\ref{subsec:lorenz-experiments}). We discretise this process with the
\gls*{statFEM}~\citep{girolami2021statistical,duffin2021statistical,duffin2022lowrank,akyildiz2022statistical},
forming the basis of the physics-informed prior on the latent space.
Stochastic additive forcing inside the \gls*{PDE} represents
additive model error, which results from possibly misspecified physics. Full
details, including the ODE case, are given in \ref{app:complete-dynamical-model}.

We assume that the model has possibly unknown coefficients
$\mLambda$ (henceforth referred to as `model parameters') and that these parameters are constant in time. On these we place the Bayesian prior $\mLambda \sim
p(\mLambda)$~\citep{stuart2010inverse}, describing our \textit{a
priori} knowledge on the model parameters before observing any
data. We also assume that $\vu_0$, the initial condition, is known up to measurement noise,
with prior $p(\vu_0)$ set accordingly. For pedagogical purposes, we
derive the discrete-time dynamic model using the \gls*{KdV} equation
as a running example, which is used in later sections as an example
\gls*{PDE}. \gls*{KdV} is given by:
\begin{equation}
  \label{eq:PDE-formulation}
  \partial_t u  + \alpha u \partial_s u  + \beta \partial_s^3u + \nu u = \dot{\xi},
  \quad \dot{\xi} \sim \mathcal{GP}(0, \delta(t-t') \cdot k(s, s')),
\end{equation}
where $u := u(s, t) \in \bR$, $\xi := \xi(s, t)$,
$s \in [0, L]$, $t \in [0,T]$,
and $\mLambda \subseteq \{\alpha, \beta, \nu\}$. Informally $\dot{\xi}$ is a
\gls*{GP}, with delta correlations in time,
and spatial correlations given by the covariance kernel
$k(\cdot, \cdot) \colon \bR \times \bR \to
\bR$~\citep{williams2006gaussian}. This is an uncertain term in the
\gls*{PDE}, representing possible model misspecification. We
assume all \gls*{GP} hyperparameters are known in this work but note that this methodology can be extended to estimate them via the marginal likelihood \cite{williams2006gaussian} or in a fully Bayesian manner \cite{lalchand2020approximate}. The
\gls*{KdV} equation is used to model nonlinear internal
waves~\citep[see, e.g., ][]{drazin1989solitons}, and describes the balance between nonlinear
advection and dispersion. Note that although the \gls*{KdV} equation
defines a scalar field, $u(s, t)$, the approach similarly holds for
vector fields.

Following \gls*{statFEM}, the equations are first spatially discretised with the \gls*{FEM} \cite{brenner2008mathematical}, then discretised in time. Equation~\ref{eq:PDE-formulation} is multiplied with a smooth test function $w(s) \in W$, where $W$ is an appropriate function space, and integrated over the domain $\Omega$ to give the weak form~\citep{thomeeGalerkinFiniteElement2006}
\[
  \langle \partial_t u, w \rangle
  + \alpha \langle u \partial_su, w \rangle
  + \beta \langle \partial_s^3 u, w \rangle
  + \nu \langle u, w \rangle
  = \langle \dot{\xi}, w \rangle,
\]
where $\langle \cdot, \cdot \rangle$ is the $L^2(\Omega)$ inner product.

The domain is discretised to give the mesh $\Omega_h$ with
vertices $\{ s_j \}_{j = 0}^{n_h}$. In this case, we take the $s_j$ to
be a uniformly spaced set of points, so that $s_j = j h$ ($h$ gives
the spacing between grid-points). On the mesh a
set of polynomial basis functions
$\{\phi_j(s)\}_{i = 1}^{n_u}$ is defined, such that
approximation to the PDE. Letting
$u_h(s, t) = \sum_{i = 1}^{n_u} u_i(t) \phi_i(s)$, the
weak form is now rewritten with these basis functions
\[
  \langle \partial_t u_h, \phi_j \rangle
  + \alpha \langle u_h \partial_s u_h, \phi_j \rangle
  + \beta \langle \partial_s^3 u_h, \phi_j \rangle
  + \nu \langle u_h, \phi_j \rangle
  =  \langle \dot{\xi}, \phi_j \rangle,
  \quad j = 1, \ldots, n_u.
\]
This gives a finite-dimensional \gls*{SDE} over the \gls*{FEM} coefficients $\vu(t) = (u_1(t),
\ldots, u_{n_u}(t))^\top$:
\[
  \mM \frac{\md \vu}{\md t}
  + \alpha \cF(\vu)
  + \beta \mA \vu  
  + \nu \mM \vu  
  = \dot{\bm{\xi}}, \quad
  \dot{\bm{\xi}}(t) \sim \NPDF(\bm{0}, \delta(t - t') \cdot \mG),
\]
where $\mM_{ij} = \langle \phi_i, \phi_j \rangle$,
$\mA_{ji} = \langle \partial_s^3 \phi_i, \phi_j \rangle$,
$\cF(\vu)_j = \langle u_h \partial_s u_h, \phi_j \rangle$,
and $\mG_{ij} = \langle \phi_i, \langle k(\cdot, \cdot), \phi_j \rangle \rangle$.
Time discretisation eventually gives the transition density
$p(\vu_n | \vu_{n-1}, \mLambda)$, for $\vu_n = \vu(n\Delta_t)$, whose
form is dependent on the discretisation used (see \ref{app:complete-dynamical-model}).

\subsection{Likelihood}
\label{subsec:likelihood}
The second component of the generative model is the \textit{likelihood} $p(\vx_n | \vu_n)$, for pseudo-data $\{\vx_n\}_{n = 1}^N$. 
The introduction of the likelihood in the model is usually necessary, as high-dimensional observations $\{\vy_n\}_{n = 1}^N$ may only be generated by some observed dimensions of $\{\vu_n\}_{n = 0}^N$. For example, perhaps it is known \textit{a priori} that only a single component of a latent coupled differential equation generates the observations (see also Section~\ref{subsec:lorenz-experiments}).
This explicit likelihood is introduced to separate the encoding process from the state-space inference; in practice we sample the pseudo-data from the variational encoding distribution $\vx_{1:N} \sim q_{\phi}(\cdot | \vy_{1:N})$, then condition on it with standard nonlinear filtering algorithms~\citep{fraccaro2017disentangled}.

The latent states $\vu_n$ are mapped at discrete times to pseudo-observations via $\vx_{n} = \mH \vu_{n} + \vr_n$, with $\vr_n \sim \NPDF(\bm{0},\mR)$. This gives the observation density $p(\vx_n | \vu_n) = \NPDF(\mH \vu_{n},\mR)$. Both the pseudo-observation operator $\mH \in \bR^{n_x \times n_u}$ and the noise covariance $\mR \in \bR^{n_x \times n_x}$ are assumed to be known in this work. See \ref{app:complete-dynamical-model} for further details on the construction of the observation model. The combination of the transition and observation densities provides the nonlinear Gaussian \gls*{SSM}:
\begin{align*}
  \text{Transition: }\quad \vu_n &= \mathcal{M}(\vu_{n-1}) + \ve_{n - 1}, \quad \ve_{n - 1} \sim \mathcal{N}(\mathbf{0}, \mQ),\\
  \text{Pseudo-observation: }\quad \vx_n &= \mH\vu_n + \vr_n, \quad \vr_n \sim \mathcal{N}(\mathbf{0}, \mR).
\end{align*}
The \gls*{ExKF}~\citep{jazwinski1970stochastic,law2015data} is used to infer the latent states given the pseudo-observations by maintaining a Gaussian approximation to the filtering posterior $p(\vu_{n} | \vx_{1:n}, \mLambda) \approx \NPDF(\vm_n, \mC_n)$, that is recursively updated. The filtering posterior $p(\vu_n|\vx_{1:n},\mLambda)$ is the distribution of the current latent state, $\vu_n$, given the observations up to and including the current time, $\vx_{1:n}$. Here we can see the recursive prediction and update steps
\begin{align*}
    \text{Prediction:}\quad & p(\vu_{n+1} | \vx_{1:n}, \mLambda) = \int p(\vu_{n} | \vx_{1:n}, \mLambda)p(\vu_{n+1}|\vu_{n}, \mLambda)\md \vu_{n}\\
    \text{Update:}\quad & p(\vu_{n+1} | \vx_{1:n+1}, \mLambda) \propto p(\vx_{n+1}|\vu_{n+1})p(\vu_{n+1} | \vx_{1:n+1}, \mLambda)
\end{align*}
Firstly the next latent state is predicted using the transition density $p(\vu_{n+1}|\vu_{n}|\mLambda)$, then the filtering posterior is updated based on the next observation using the observation density $p(\vx_{n+1}|\vu_{n+1})$. For a linear Gaussian \gls*{SSM}, the Kalman filter computes these steps exactly. However when the \gls*{SSM} is nonlinear these steps can be computed approximately using an appropriate nonlinear filter, e.g. the \gls*{ExKF}. Details on the prediction and update steps of the \gls*{ExKF} are provided in \ref{app:complete-dynamical-model}. 

\subsection{Decoder}
\label{subsec:decoder}
The last component of our generative model is the decoder $p_\theta(\vy_n | \vx_n)$, which describes the unknown mapping between the pseudo-observations $\vx_n$, and the observed data $\vy_n$. The decoding of latents to data should model as closely as possible the true data generation process. Prior knowledge about this process can be used to select an appropriate $p_{\theta}(\vy_n|\vx_n)$. No temporal structure is assumed on $\theta$, so the decoder is shared across all times $p_{\theta}(\vy_{1:N} | \vx_{1:N}) = \prod_{n = 1}^N p_{\theta}(\vy_n | \vx_n)$. The function $f_{\theta}\colon \mathbb{R}^{n_x} \rightarrow \mathbb{R}^{n_y}$ maps the latent encoding to the data reconstruction $\hat{\vy}_n = f_{\theta}(\vx_n)$, and for a simple Gaussian probabilistic decoder this reconstruction is the mean, and a Gaussian reconstruction error is assumed, i.e. $\vy_n \sim \mathcal{N}(f_{\theta}(\vx), \sigma^2 \mI)$. For more details on specific architectures defining $f_{\theta}$ see \ref{app:numerics-and-architectures}.

\begin{figure}[ht!]
    \centering
    \includegraphics[width=0.9\textwidth]{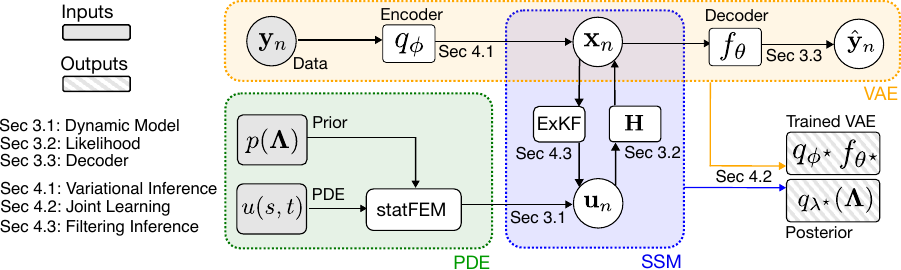}
    \caption{Flow diagram describing connections between the specified \acrfull*{PDE}, latent \acrfull*{SSM} and \acrfull*{VAE}. Inputs to the method include the unstructured data $\vy_{1:N}$, the specified \gls*{PDE} $u(x, s)$ and the model parameter prior (solid grey fill). Outputs are the trained autoencoder parameters $\left\{\phi^\star, \theta^\star\right\}$, and the model parameter posterior distribution parameters $\lambda^\star$ (striped grey fill). Labels are included referencing the relevant sections.}
    \label{fig:flow-diagram}
\end{figure}

\section{Inference}
\label{sec:inference}

In the previous section we introduced the probabilistic model, describing the joint distribution of the data $\vy$, pseudo-observations $\vx$ and latent states $\vu$. This section develops the methodology for performing inference on the latent states based on this probabilistic model, i.e. conditional on the data, what is the posterior distribution of the pseudo-observations and latent states? The inference can be broken down into two parts, the \gls*{VAE} which encodes high-dimensional data into the space of pseudo-observations, approximating $p(\vx_n|\vy_n)$, and the \gls*{ExKF} for inferring the latent states given the pseudo-observations, approximating $p(\vu_n|\vx_{1:n}, \mLambda)$. Figure \ref{fig:flow-diagram} visualises how the \gls*{VAE} and the latent \gls*{SSM} interact, and includes references to the relevant subsections where the parts of the model are described.

Section \ref{subsec:variational-inference} describes the inference of the pseudo-observations given the data; this posterior is estimated using a variational approximation, constructed via a \gls*{VAE}. Section \ref{subsec:joint-learning} describes how we infer the model parameters using a \gls*{VB} approximation to the model parameter posterior. Finally we detail the filtering inference in Section \ref{subsec:filtering-inference}, where the \gls*{ExKF} is used to approximate the filtering posterior and make predictions of the latent states at future times.

\subsection{Variational Inference}
\label{subsec:variational-inference}
In this section, we introduce the variational family and the \gls*{ELBO}. Denote by $q(\vu_{1:N}, \vx_{1:N}, \mLambda| \vy_{1:N})$ the variational posterior, which, similar to \citet{fraccaro2017disentangled}, factorises as
\begin{equation}\label{eq:variational-factorisation}
  q(\vu_{1:N}, \vx_{1:N}, \mLambda| \vy_{1:N})
  = q(\vu_{1:N} | \vx_{1:N}, \mLambda) q_{\phi}(\vx_{1:N}|\vy_{1:N}) q_\lambda(\mLambda).
\end{equation}
Note here that we do not make a variational approximation  $q(\vu_{1:N}| \vx_{1:N}, \mLambda)$; this is taken to be the exact posterior $p(\vu_{1:N} | \vx_{1:N}, \mLambda)$. We derive the \gls*{ELBO} to be~(see also \ref{app:complete-variational-derivation})
\begin{align}\label{eq:ELBO}
  \log p(\vy_{1:N}) &\geq  \int \log \left[  \frac{p(\vu_{1:N}, \vx_{1:N}, \mLambda , \vy_{1:N})}{q(\vu_{1:N}, \vx_{1:N}, \mLambda | \vy_{1:N})}\right]
                      q(\vu_{1:N}, \vx_{1:N}, \mLambda | \vy_{1:N})  \md \vx_{1:N} \md \vu_{1:N} \md \mLambda \nonumber \\
                    &= \mathbb{E}_{q_{\phi}} \left[
                        \log \frac{p_{\theta}(\vy_{1:N} | \vx_{1:N})}{q_{\phi}(\vx_{1:N}|\vy_{1:N})}
                        +
                      \mathbb{E}_{q_{\lambda}}
                      \left[
                      \log p(\vx_{1:N}|\mLambda)
                        + \log \frac{p(\mLambda)}{q_{\lambda}(\mLambda)}
                        \right]
                      \right].
\end{align}
Typically this expectation is not analytically tractable and \gls{MC} is used to compute an approximation. Estimating $\log p(\vx_{1:N} | \mLambda)$ requires marginalising over $\vu_{1:N}$, \textit{the latent physics process}. As detailed above, the \gls*{ExKF} recursively computes a Gaussian approximation to the filtering posterior $p(\vu_n|\vx_{1:n}, \mLambda) \approx \NPDF(\vm_n, \mC_n)$. 
We note however that this can also be realised with other nonlinear filters, e.g., ensemble Kalman filters~\citep{chen2022autodifferentiable} or particle filters~\citep{corenflos2021differentiable}.
The factorisation of the pseudo-observation marginal likelihood, $p(\vx_{1:N} | \mLambda) = p(\vx_1 | \mLambda) \prod_{n = 2}^N p(\vx_{n} | \vx_{1:n - 1}, \mLambda)$, enables computation, as the filter can compute $p(\vx_{n} | \vx_{1:n - 1}, \mLambda)$, at each prediction step, via $p(\vx_{n} | \vx_{1:n - 1}, \mLambda) = \int p(\vx_n | \vu_n, \mLambda) p(\vu_n | \vx_{1:n - 1}, \mLambda) \, \md \vu_n$.

Encoder parameters $\phi$ are shared between variational distributions $\left\{ q_{\phi}(\vx_n|\vy_n) \right\}_{n=1}^N$ to give an amortized approach~\citep{kingma2019introduction}. Unless otherwise specified, for each $n$ the encoding has the form $q_\phi(\vx_n | \vy_n) = \NPDF(\mu_\phi(\vy_n), \sigma_\phi(\vy_n))$. Functions $\mu_\phi(\vy_n) \colon \bR^{n_y} \to \bR^{n_x}$ and $\log\sigma_\phi(\vy_n) \colon \bR^{n_y} \to \bR^{n_x}$ are \glspl*{NN}, with parameters $\phi$ to be learnt. Parameter updates are performed using the Adam optimiser \cite{kingma2017adam}.
Practical implementation details and specific encoding architectures are given in \ref{app:numerics-and-architectures}. As for $q_\lambda$, the form of the variational approximation is set based on the parameters of interest, see Table~\ref{tab:variational} for a few candidate distributions. 

\subsection{Joint Learning}
\label{subsec:joint-learning}

The inverse problem we aim to solve is the joint estimation of the filtering posterior, $p(\vu_n | \vy_{1:n})$, and the model parameters $\mLambda$. 
Here, we assume the latent transition distribution, $p(\vu_n | \vu_{n-1}, \mLambda)$ is known and can be constructed from the PDE via \gls{statFEM}, and treat $\mLambda$ as a parameter we wish to estimate via the variational approximation in Eq.~\ref{eq:variational-factorisation}. 
The prior distribution $p(\mLambda)$ in effect places a prior over a family of partial differential equations, whose posterior is targeted by learning the parameters of the variational approximation $q_{\lambda}(\mLambda)$. 
These distribution parameters $\lambda$ are learned via gradient ascent of the \gls{ELBO}, using MC samples from the variational posterior $\mLambda^{(i)} \sim q_{\lambda}(\cdot)$.

\begin{algorithm}[!ht]
    \DontPrintSemicolon
      \KwInput{training data $\vy_{1:N}$, \gls*{SSM} $\left\{ \mathcal{M}, \mH, \mQ, \mR \right\}$, model parameter prior $p(\mLambda)$, initial parameters $\left\{ \phi_0, \theta_0, \lambda_0 \right\}$, step number $K$}
      \For{$k = 0, \dots, K-1$}
            { 
                $
                    \vx_{1:N} \sim q_{\phi_k}(\cdot | \vy_{1:N})
                $
                \Comment*[r]{Sample pseudo-observations}
                $
                    \mLambda \sim q_{\lambda_k}(\cdot)\;
                $
                \Comment*[r]{Sample model parameters}
                $
                    \mathcal{L}_k = \log p_{\theta_k}(\vy_{1:N} | \vx_{1:N}) - \log q_{\phi_k}(\vx_{1:N}|\vy_{1:N})
                          + \log p(\vx_{1:N}|\mLambda) - \kl{q_{\lambda_k}}{p}\; 
                $
                \Comment*[r]{Calculate ELBO}
                $
                \left( \phi_{k+1}, \theta_{k+1}, \lambda_{k+1} \right) = \text{Adam}\left( \nabla \mathcal{L}_k, \phi_{k}, \theta_{k}, \lambda_{k} \right)\;
                $
                \Comment*[r]{Adam update step}
            }
        \KwRet{ $\left( \phi_K, \theta_K, \lambda_K \right)$ }
    \caption{\gls*{PIDVAE}: joint learning of autoencoder and model parameter distribution}
    \label{algo:phi-dvae}
\end{algorithm}

This relies on the KL-divergence between the prior and variational posterior being available analytically, which is the case for some choices of prior and variational posterior. We also require a closed form reparameterisation for sampling from the posterior to allow for gradient computation. Table \ref{tab:variational} details the relevant formulae for a few popular univariate distributions. Note that results are available for multivariate cases, e.g. see~\citep{williams2006gaussian} for multivariate Gaussian reparametrisation via the Cholesky decomposition, and analytic KL-divergence between two multivariate Gaussians. Pseudo-code for the joint learning is provided in Algorithm~\ref{algo:phi-dvae}.

\begin{table}[ht]
\centering
\begin{tabular}{@{}llllll@{}}
\toprule
\textbf{Distribution} & $\lambda$            & \textbf{Support}         & \textbf{PDF}                                                                          & \textbf{KL-Divergence}                                                                                         & \textbf{Reparameterise}   \\ \midrule
Gaussian              & $\left\{ \mu, \sigma \right\}$ & $x\in \mathbb{R}$ & $\frac{1}{\sigma \sqrt{2\pi}} \exp\left( -\frac{(x-\mu)^2}{2\sigma^2} \right)$        & $\frac{\mu_2}{2} - \frac{\mu_1}{2} - \frac{1}{2} + \frac{\sigma_1^2 + (\mu_1 - \mu_2)^2}{2\sigma_2^2}$               & $\mu + \sigma * Z$       \\
Exponential           & $\left\{ \lambda \right\}$     & $x\in \mathbb{R}^+$       & $\lambda \exp(-\lambda x)$                                                            & $\log\left(\frac{\lambda_1}{\lambda_2}\right) + \frac{\lambda_2}{\lambda_1} - 1$                               & $-\log(U)/\lambda$       \\
Log-normal             & $\left\{ \mu, \sigma \right\}$ & $x\in \mathbb{R}^+$       & $\frac{1}{x\sigma \sqrt{2\pi}} \exp\left( -\frac{(\log(x)-\mu)^2}{2\sigma^2} \right)$ & $\log\left(\frac{\sigma_2}{\sigma_1}\right) + \frac{(\mu_1 - \mu_2)^2 + \sigma_1^2 - \sigma_2^2}{2\sigma_2^2}$ & $\exp(\mu + \sigma * Z)$ \\ \bottomrule
\end{tabular}
\caption{Variational distributions for parameters, $Z \sim \mathcal{N}(0,1), U \sim \mathcal{U}[0,1]$.}
\label{tab:variational}
\end{table}

\subsection{Filtering Inference}
\label{subsec:filtering-inference}

Once the autoencoder parameters $\left\{ \phi, \theta \right\}$ have been learned jointly with the variational parameters $\lambda$, we can assess the performance of the \gls{PIDVAE} algorithm by comparing the filtering posterior $p(\vu_{n}|\vy_{1:n})$, to the true latent solution $\vu(n\Delta_t)$ that generated the data. This assesses the ability of both the model to provide meaningful encodings via the autoencoder, and the extended Kalman filter to give latent solutions that match the true data generating solution. This distribution can be approximated via \gls*{MC} samples from the variational autoencoder, marginalising over the latent encodings $\vx_{1:N}$ and the model parameters $\mLambda$. We can approximate the filtering marginal distribution at a time $n$:
\begin{align}\label{eq:filter-marginal}
  p(\vu_n|\vy_{1:n})
  &\approx \int p(\vu_n| \vx_{1:n}, \mLambda) 
  q_\lambda(\mLambda) q_{\phi}(\vx_{1:n}|\vy_{1:n}) \, \md \mLambda \md \vx_{1:n} \nonumber\\
  &\approx \frac{1}{M} \sum_{i=1}^{M} p(\vu_n|\vx_{1:n}^{(i)}, \mLambda^{(i)}), 
    \quad 
    \vx_{1:n}^{(i)} \sim q_{\phi}(\cdot|\vy_{1:n}), \;
    \mLambda^{(i)} \sim q_{\lambda}(\cdot).
\end{align}
The resulting distribution is a mixture of Gaussians over the latent state, with each
$p(\vu_n|\vx_{1:n}^{(i)}, \mLambda^{(i)}) = \mathcal{N}(\vm_n^{(i)}, \mC_n^{(i)})$.

The model parameter inference can be assessed by comparing the variational posterior $q_{\lambda}(\mLambda)$ to the prior distribution $p(\mLambda)$ and the true coefficient value that generated the data.

Finally we can compare the performance of the joint estimation by marginalising over the model parameters $\mLambda$, and comparing this to a model that learned the autoencoder parameters conditioning only on samples from the prior $p(\mLambda)$. 
We denote this trained prior model $\left\{ \phi, \theta \right\}$, and denote the jointly trained model $\left\{ \phi^{\star}, \theta^{\star}, \lambda^{\star} \right\}$.
This should isolate the impact of learning a variational approximation to the model parameters $\mLambda$ on the resulting latent inference and predictive power. 
Ideally, the learned posterior should be more concentrated around the ground truth value of the model parameter, giving predictions of latent states that are closer to the true latent process. 
For prediction, the aim will be to infer the values of the latent process past the final time of the training dataset; these predictions are made by iterating only the prediction step of the Kalman filter ahead in time from the final mean and covariance. 
We will refer to these as \textit{rollout predictions}, as in essence we are `rolling out' the filter to provide future predictions without incorporating observations.

For the rollout, the target distribution in the latent state a number of time steps $\tau$ ahead of step $n$, i.e. $p(\vu_{n+\tau} | \vy_{1:n})$. Consider the distribution over the latents a single time step ahead,
 
\begin{align} 
    p(\vu_{n+1} | \vy_{1:n}) &\approx \int p(\vu_{n+1}| \vu_{n}, \mLambda)p(\vu_n| \vx_{1:n}, \mLambda) q_\lambda(\mLambda) q_{\phi}(\vx_{1:n}|\vy_{1:n}) \md\vu_{n} \md\vx_{1:n} \md\mLambda \nonumber \\ 
    &= \int \underbrace{\left[ \int p(\vu_{n+1}| \vu_{n}, \mLambda)p(\vu_n| \vx_{1:n}, \mLambda) \md\vu_{n} \right]}_{\text{Kalman Filter Prediction Step}} q_\lambda(\mLambda) q_{\phi}(\vx_{1:n}|\vy_{1:n}) \md\vx_{1:n} \md\mLambda \nonumber \\
    &= \int p(\vu_{n+1}| \vx_{1:n}, \mLambda) q_\lambda(\mLambda) q_{\phi}(\vx_{1:n}|\vy_{1:n}) \md\vx_{1:n} \md\mLambda \nonumber \\ 
    &\approx \frac{1}{M} \sum_{i=1}^{M} p(\vu_{n+1}|\vx_{1:n}^{(i)}, \mLambda^{(i)}), 
    \quad 
    \vx_{1:n}^{(i)} \sim q_{\phi}(\cdot|\vy_{1:n}), \;
    \mLambda^{(i)} \sim q_{\lambda}(\cdot). \label{eq:filter-predict-one-step}
\end{align}

Here, the mixture of Gaussians distribution is constructed with the one-step-ahead predictive distributions $p(\vu_{n+1}|\vx_{1:n}^{(i)}, \mLambda^{(i)}) = \mathcal{N}(\hat{\vm}_{n+1}^{(i)}, \hat{\mC}_{n+1}^{(i)})$. This step can be iterated on the mean and covariance to give predictions $\tau$ steps ahead, i.e. the distribution $p(\vu_{n+\tau}|\vx_{1:n}^{(i)}, \mLambda^{(i)})$.

\section{Experiments}
\label{sec:experiments}
In this section, we aim to simulate the following scenario where a practitioner with an experimental setup recording unstructured data generated by a physical process. There is a mechanistic model the practitioner believes to be the underlying physical process, with some unknown model parameters described through a prior distribution. Given this setup, our model can be used to embed this unstructured data into the latent space described by the physical model. The resulting output is a posterior distribution of the latent states given the unstructured data (given by embedding the data with the trained \gls*{VAE} and filtering with the \gls*{ExKF}, giving $p(\vu_n|\vy_{1:n})$), and an updated estimate of the model parameters (in the form of a variational posterior $q_{\lambda}(\mLambda)$).
We provide three experimental examples, each with different latent differential equations:
\begin{itemize}
    \item \textbf{Advection equation:} First, we introduce a \textit{linear} \gls*{PDE} example, in the form of the advection equation with video data as unstructured observations. Due to the linearity of the latent space, a vanilla Kalman filter can be used for filtering pseudo-observations. Therefore, this example provides a valid comparison to to another \gls*{DVAE} method, the \gls*{KVAE}, which uses a vanilla Kalman filter and learns a general linear Gaussian \gls*{SSM}. 
    \item \textbf{Lorenz-63:} The second example extends this to the Lorenz-63 system of \textit{nonlinear} ordinary differential equations, where we demonstrate the \gls*{ExKF} for nonlinear filtering. In this example, we use velocity field measurements as the unstructured data. Experiments consider the joint estimation of a model parameter and inference of unobserved latent states.
    \item \textbf{\gls*{KdV} equation:} Finally, our most complex example is the \textit{nonlinear} and \textit{partial} differential equation: the \gls*{KdV} equation, with video data observations. Here, both \gls{statFEM} and \gls*{ExKF} are employed. We demonstrate the ability of \gls*{PIDVAE} to adjust for model parameter misspecification and improve future latent prediction, purely using information extracted from video data.
\end{itemize}

We choose to affect our data with noise, to reflect the manner in which this data would be collected. For velocity field data, additive Gaussian noise is applied to simulate measurement errors, and for video data we include salt-and-pepper noise~\citep{boncelet2009image} to simulate noise for image frames.

\subsection{Advection PDE}
\label{sec:advection-results}
As our first example, we consider the advection equation with periodic boundary
conditions. In this example, we derive the transition
density $p(\vu_n | \vu_{n - 1}, \mLambda)$ from a \gls*{statFEM} discretisation
of a stochastic advection equation:
\begin{equation}
  \partial_t u + c \, \partial_s u = \dot{\xi}, \quad
  \dot{\xi} \sim \mathcal{GP}(0, \delta(t - t') \cdot k(s, s')), \label{eq:advection}
\end{equation}
where $u := u(s, t)$, $\xi := \xi(s, t)$, $s\in [0, 1]$, $t\in [0,40]$, and $u(s, t) = u(s + 1, t)$.
Recall that, as in Section~\ref{subsec:likelihood}, the FEM coefficients
$\vu_n = (u_1(n \Delta_t), \ldots, u_{n_u}(n \Delta_t))$ are the latent
variables. These are related to the discretised solution
via $u_h(s, n \Delta_t) = \sum_{i = 1}^{n_u} u_i(n \Delta_t) \phi_i(s)$
(see~\ref{app:numerics-and-architectures}
and~\ref{app:complete-dynamical-model}). 

Video data $\vy_{1:N}$ is generated from the
deterministic version of~\eqref{eq:advection} (i.e.~\eqref{eq:advection}
with $\xi \equiv 0$). Since the solution of this equation is available analytically, we use this for data generation. Denoting the initial condition $f(s) = u(s, 0)$, the solution to ~\eqref{eq:advection} at a time $t$ is given by $u(s, t) = f(s - ct)$. 
This trajectory is evaluated at discrete time points $\left\{n\Delta_t\right\}_{n=1}^{N}$ and the solutions $u(s, n\Delta_t)$ are imposed onto a 2D grid. 
On the grid, pixels below $u(s, n\Delta_t)$ are lit-up in a binary fashion, along with salt-and-pepper noise.
In this experiment we use fixed parameters, setting the wavespeed $c = 0.5$. 
We set the decoder as $p_\theta(\vy_n | \vx_n) = \mathrm{Bern}(f_\theta(\vx_n))$ and the encoder as $q_\phi(\vx_n | \vy_n) = \NPDF(\mu_\phi(\vy_n), \sigma_\phi(\vy_n))$. As previous, see \ref{app:numerics-and-architectures} for full details.

\begin{figure}[ht!]
    \centering
    \includegraphics[width=\textwidth]{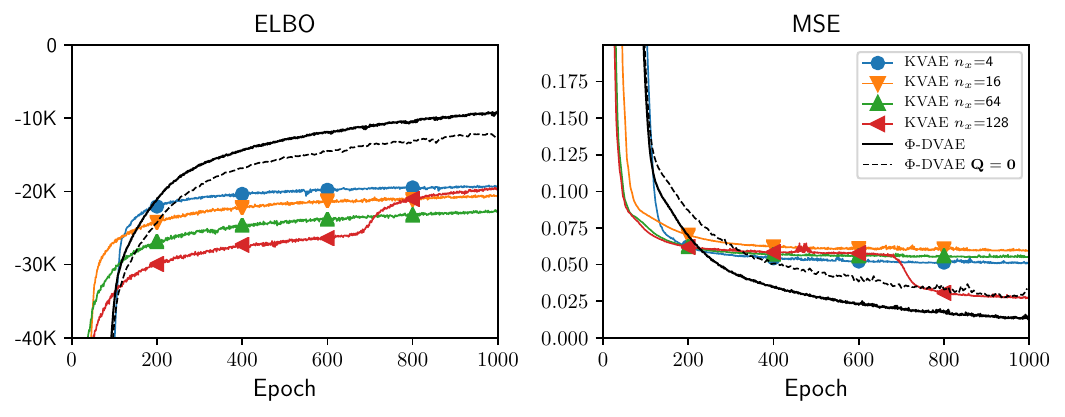}
    \caption{Comparison of \gls*{KVAE} and \gls*{PIDVAE} for the advection equation.}
    \label{fig:advection-results}
\end{figure}

Due to linearity of the underlying dynamical system, we compare the \gls*{PIDVAE} to the \gls*{KVAE} for a set of video data generated from the advection equation, for various dimensions of the \gls*{KVAE} latent space. 
Specifying a particular form of \gls*{PDE} dynamics on the latent states increases the inductive bias imposed on the latent space, and should provide better reconstructions and higher likelihood of the data. We choose two metrics to evaluate this: the \gls*{ELBO} given by $\log p_{\theta}(\vy_{1:N}|\vx_{1:N}) - \log q_{\phi}(\vx_{1:N}|\vy_{1:N}) + \log p(\vx_{1:N})$, and the normalised \gls*{MSE} of reconstruction compared to the noise-free video frames, $\vy_{1:N}^{\text{true}}$. The normalised \gls*{MSE} is given by $\Vert \vy_{1:N}^{\text{true}} - \hat{\vy}_{1:N} \Vert_{2}^{2} / \Vert \vy_{1:N}^{\text{true}} \Vert_{2}^{2}$. 
We compare our method to \gls*{KVAE} for the linear advection example for varying latent dimension of the \gls*{KVAE}, and the results are plotted in Fig.~\ref{fig:advection-results}. We also include the results for our method but with deterministic latent transitions ($\mQ=\mathbf{0}$) to assess the impact of accounting for misspecification through the use of \gls*{statFEM}.

From Fig.~\ref{fig:advection-results}, \gls*{PIDVAE} outperforms each \gls*{KVAE} model, with a larger \gls*{ELBO} and smaller normalised \gls*{MSE} during training. The \gls*{MSE} for the \gls*{PIDVAE} model is 1.26\%, compared to the best \gls*{KVAE} model ($n_x = 128$) with 2.71\%, where roughly a 5\% normalised \gls*{MSE} is considered a good reconstruction. The \gls*{KVAE} models performed similarly in terms of \gls{ELBO}, with dimension 4 and 128 the best with -19273 and -19659 respectively, but are outperformed by both \gls*{PIDVAE} models. We also note that including the process noise matrix $\mQ$ for the \gls*{statFEM} model which accounts for misspecification outperforms the model that evolves the latent states deterministically.

\subsection{Lorenz-63}
\label{subsec:lorenz-experiments}

In our next example, the latent dynamical model $p(\vu_n | \vu_{n - 1}, \mLambda)$
is given by an Euler-Maruyama discretisation~\citep{kloedenpeter1992numerical} of the stochastic Lorenz-63 system,
\begin{align}
  \label{eq:stochastic_lorenz_63_ode}
  \md {u}_1 &= (-\sigma u_{1} + \sigma u_{2}) \md t + \md w_1, \nonumber \\
  \md {u}_2 &= (-u_{1} u_{3} + r u_{1} - u_{2})  \md t + \md w_2, \\
  \md {u}_3 &= (u_{1} u_{2} - bu_{3}) \md t + \md w_3, \nonumber
\end{align}
where $\vu(t) := [u_1(t), u_2(t), u_3(t)]^\top$, $t \in [0, 6]$, $\vu_n = [u_1(n\Delta_t), u_2(n\Delta_t), u_3(n\Delta_t)]$,
$\mLambda \subseteq \left\{\sigma, r, b\right\}$, and
$w_1$, $w_2$, and $w_3$ are independent Brownian
motion processes~\citep{oksendal2003stochastic}.

\begin{figure}[ht!]
    \centering
    \includegraphics[width=\textwidth]{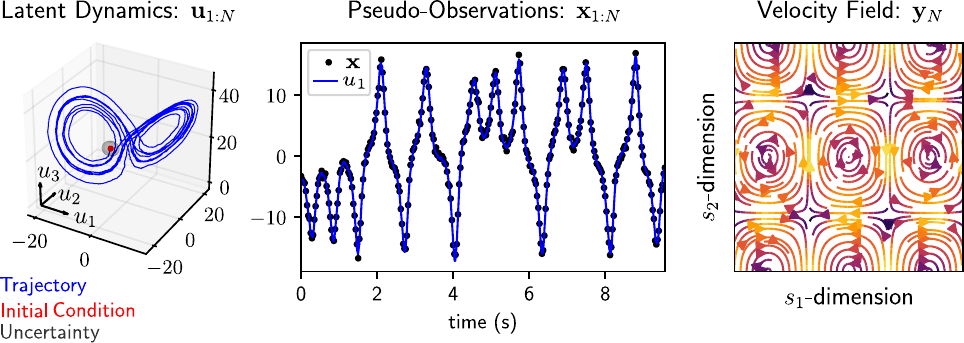}
    \caption{Lorenz-63: latent states $\vu_{1:N}$, pseudo-observations $\vx_{1:N}$, and velocity field $\vy_N$.}
    \label{fig:lorenz-vis}
\end{figure}

We observe synthetic 2D velocity fields, $\vy_{1:N}$ of convective
fluid flow and we use our method to embed these synthetic data into the
stochastic Lorenz-63 system. The Lorenz-63 system is related to
the velocity fields through a truncated spectral expansion. In brief, it is assumed that the velocity fields have
no vertical velocity, so the 3D velocity field is realised in 2D.
The velocity field can be described by the stream function
$\psi := \psi(s_1, s_2, t)$, where $s_1$ and $s_2$ are the
spatial coordinates of variation, respectively, and thus
$\mathbf{y}(t) = \left( -\partial_{s_2} \psi, 0, \partial_{s_1} \psi \right)$.
A truncated spectral approximation and a transform to non-dimensional equations
yields $\psi(s_1, s_2, t) \propto u_1(t) \sin(\pi s_1/l)\sin(\pi s_2/d)$,
where $u_1(t)$ is governed by the Lorenz-63 ODE (i.e.,
\eqref{eq:stochastic_lorenz_63_ode} with $w_i \equiv 0$).
For full details we refer to \ref{app:numerics-and-architectures} and \ref{app:complete-dynamical-model}. The Lorenz-63 system is classical system,  popularised in~\citet{lorenz1963deterministic} through its characteristic ``deterministic nonperiodic flow'': chaotic dynamics. 

To generate the synthetic data
$\vy_{1:N}$, we generate a trajectory $\vu_{1:N}^{\mathrm{true}}$ from
the deterministic version of~\ref{eq:stochastic_lorenz_63_ode}, at discrete timepoints $n \Delta_t$, and 
use the generated $u_{1,n}^{\mathrm{true}}$ to compute the two-dimensional
velocity field, $\vy_n$, via $\psi(s_1, s_2, t)$. This corresponds to
having a middle layer $x_n = u_{1,n} + w_n$ where 
$w_n \sim \mathcal{N}(0, R^2)$, with likelihood
$p(x_n | \vu_n) = \mathcal{N}(\vh^\top\vu_n, R^2)$ where $\vh = [1, 0, 0]^\top$.
The synthetic data of $\vu^{\mathrm{true}}(t)$, $\vx_{1:N}$, and $\vy_{N}$ are visualised in
Figure~\ref{fig:lorenz-vis}; the full trajectory $\vu^{\mathrm{true}}(t)$ is
shown in 3D, and the velocity field $\vy_n$ is shown in 2D, for a single $n$.
The decoding is assumed to be of the form $p_{\vw}(\vy_n | x_n) =
\NPDF(\vw x_n, \eta^2 \mI)$, with unknown coefficients
$\vw \in \mathbb{R}^{n_y}$ which are to be learned from the data. The variational encoding $q_{\vw}(x_n|\vy_n)$ is determined via a pseudo-inverse, as detailed in \ref{app:numerics-and-architectures}, along with the relevant hyperparameters and numerical details.

\subsection*{Lorenz-63 Results}
The coefficients used to generate the Lorenz-63 data are $\left\{ \sigma, b, r\right\} = \left\{ 10, 8/3, 28 \right\}$, and it is assumed both $r$ and $b$ are known, whereas $\sigma$ is not. 
A poorly specified prior $p(\mLambda) = \mathcal{N}(30,5^2)$ is placed over this unknown parameter $\mLambda = \sigma$, in order to test the model's ability to correct this misspecification through the adjustment of the variational posterior.\\
For this experiment, we first train the linear autoencoder parameters while sampling the model parameter from the prior distribution. This pre-training step is to allow sensible VAE embeddings to be learned before adjusting the variational posterior over model parameters \cite{fraccaro2017disentangled}.  
The parameters of the variational distribution $\lambda = \left\{ \mu_\lambda, \sigma_\lambda \right\}$ are then jointly estimated with the autoencoder parameters to train the complete model, $\left\{ \mathbf{w}^{\star}, \lambda^{\star} \right\}$. See Algorithm~\ref{algo:phi-dvae}.

\begin{figure}[ht]
    \centering
    \includegraphics[width=\textwidth]{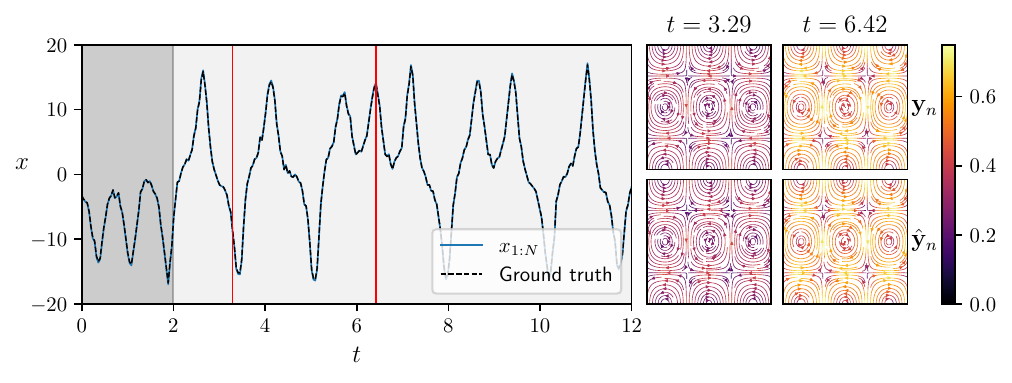}
    \caption{\textit{Left:} Ground truth pseudo-observations (dashed) compared to encoding $\vx_{1:N} \sim q_{\phi^{\star}}(\cdot | \vy_{1:N})$ (solid). Training data indicated by grey-fill ($t=\left[ 0, 2\right)$), red lines indicate reconstruction times. \textit{Right:} Observed data $\vy_n$ plotted alongside reconstructions $\hat{\vy}_{n}$, with the velocity field shown as a streamplot where the color indicates the speed of the flow.}
    \label{fig:lorenz-encoding}
\end{figure}

\begin{figure}[t!]
  \centering
  \begin{subfigure}[t]{0.28\textwidth}
    \centering
    \includegraphics[width=\textwidth]{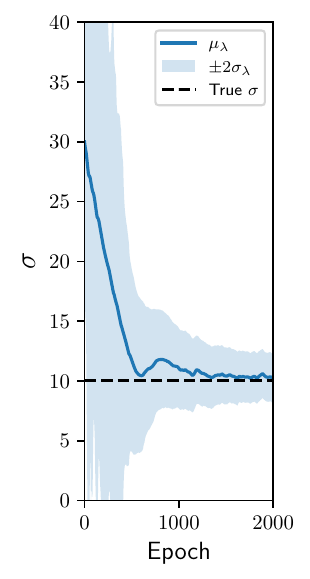}
    \caption{}
    \label{fig:lorenz-variational-posterior}
  \end{subfigure}
  \begin{subfigure}[t]{0.70\textwidth}
    \centering 
    \includegraphics[width=\textwidth]{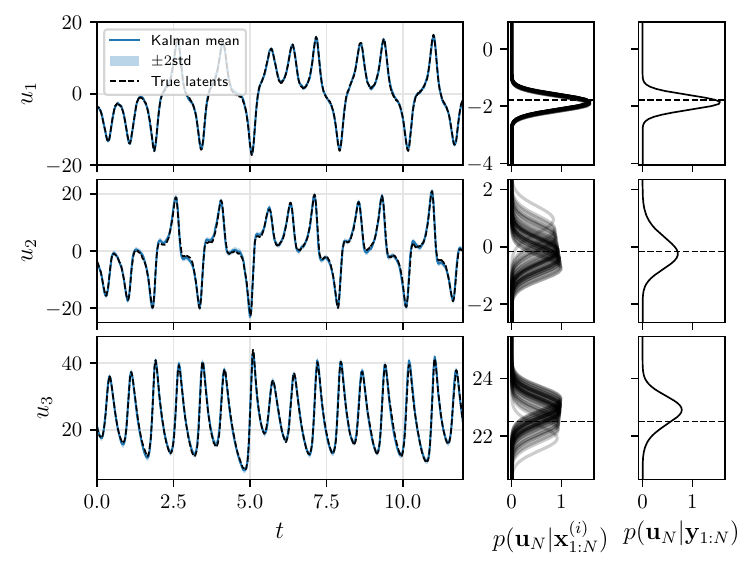}
    \caption{
    }
    \label{fig:lorenz-inference}
  \end{subfigure}
  \caption{
    Inference results for \gls*{PIDVAE} jointly trained on Lorenz-63 system. 
    \text{(a)} 
    Trace plot of parameter variational distribution $q_{\lambda}(\mLambda) = \mathcal{N}(\mu_{\lambda}, \sigma_{\lambda}^2)$, with mean (blue) and $\pm 2$ standard deviations (blue fill). Initialised at poorly specified prior, $p(\mLambda) = \mathcal{N}(30, 5^2)$. True value $\sigma = 10$ (dashed).
    \text{(b)} 
    \textit{First column:} Filtering inference for latent states $\vu_{1:N}$, where the filtering distribution $p(\vu_n | \vx_{1:n})$ is plotted with the ground truth $\vu^{\text{true}}_n$.
    \textit{Second column:} Final-time filtering distribution $p(\vu_N | \vx_{1:N}^{(i)})$, for multiple samples $\vx_{1:N}^{(i)}, i=1,\dots,50$.
    \textit{Third column:} Mixture of Gaussians distribution, $p(\vu_N | \vy_{1:N})$, calculated via Eq.~\ref{eq:filter-marginal}. True latent values $\vu_{N}^{\text{true}}$ shown as horizontal dashed lines.
    }
  \label{fig:lorenz-results}
\end{figure}

\begin{figure}[ht!]
\centering
  \includegraphics[width=\textwidth]{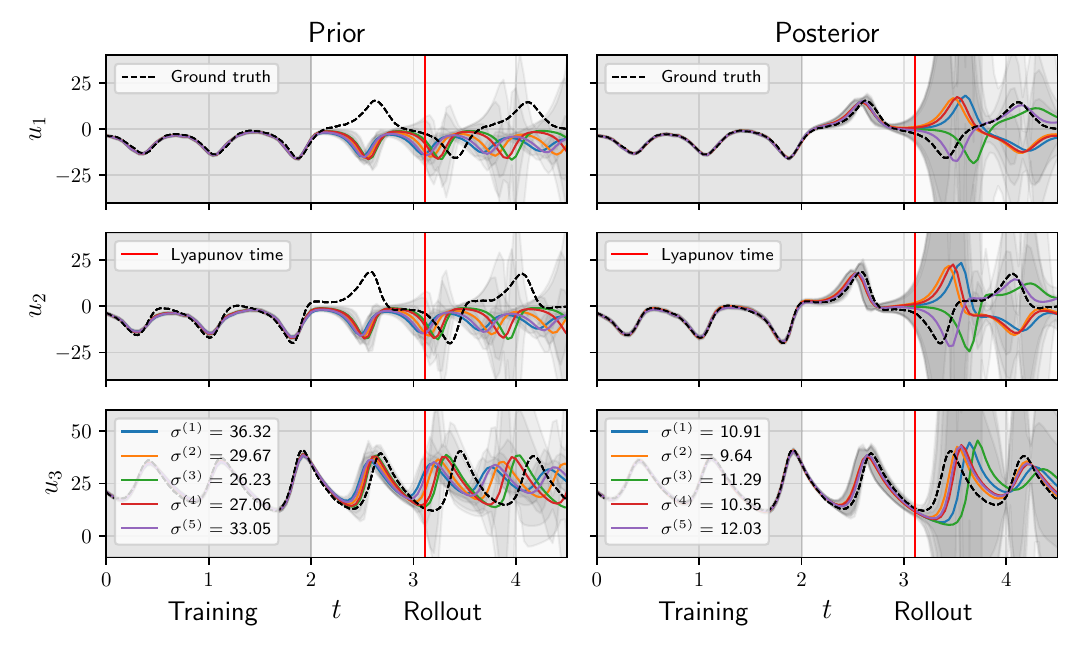}
  \caption{
    Training time indicated with grey-fill $t=\left[0,2\right)$, and rollout time from $t=\left[2,4.5\right]$. Filtering distributions $p(\vu_n | \vx_{1:n}^{(i)}, \mLambda^{(i)}) = \mathcal{N}(\vm_n^{(i)}, \mC_n^{(i)})$ visualised with $\vm_n^{(i)}$ (solid) and $\pm 2$ standard deviations computed from $\mC_n^{(i)}$ (grey-bands), alongside the true latents $\vu_{1:N}^{\text{true}}$ (dashed).
    \textit{First column:} samples generated by the prior model, $\vx_{1:N}^{(i)} \sim q_{\phi}(\cdot | \vy_{1:N})$, $\mLambda^{(i)} \sim p(\cdot)$.
    \textit{Second column:} samples generated from the posterior model, $\vx_{1:N}^{(i)} \sim q_{\phi^{\star}}(\cdot | \vy_{1:N})$, $\mLambda^{(i)} \sim q_{\lambda^{\star}}(\cdot)$.\\
  }
  \label{fig:lorenz-samples}
\end{figure}

The training data consists of the first 50 velocity field measurements. The latent states that generated these measurements come from only one `wing' of the characteristic butterfly-shaped Lorenz-63 attractor. 
This assesses the model's ability to extrapolate, and encode velocity measurements unseen during training that are generated from the other `wing' of the attractor. 
The full dataset $\vy_{1:N}$ is encoded using the trained variational autoencoder, and a sample from the variational posterior is compared to the ground truth pseudo-observations $\vx_{1:N}^{\text{true}}$ in Fig.~\ref{fig:lorenz-encoding}. The sampled encoding is close to both the training and test ground truth pseudo-observations.
Results for the estimation of parameter $\sigma$ during the joint training are given in Fig.~\ref{fig:lorenz-variational-posterior}. The variational distribution is initialised at the prior $q_{\lambda_0}(\mLambda) = \mathcal{N}(30, 5^2)$, and over the joint training period the mean converges on the true value $\sigma=10$, and the variance reduces and appears to level off. Fig.~\ref{fig:lorenz-inference} shows the latent inference for the full dataset, the first column shows the distribution $p(\vu_n | \vx_{1:n}, \mLambda)$, with sampled $\vx_{1:n} \sim q_{\phi^{\star}}(\cdot | \vy_{1:N})$ and $\mLambda \sim q_{\lambda^{\star}}(\cdot)$ which shows good correspondence to the true latent values. The third column shows the mixture of Gaussian approximation to $p(\vu_n | \vy_{1:n})$ (see Eq.~\ref{eq:filter-marginal}), with the second column visualising the individual mixture components. As expected, the observed latent dimension $u_1$ has a smaller variance than the unobserved components $u_2, u_3$, with low bias seen for all three dimensions. 
Fig.~\ref{fig:lorenz-samples} shows the results of the predictive rollout. In this scenario, the training data is encoded from $t=0$ to $t=2$, and the latent states are predicted ahead in time by iterating Eq.~\ref{eq:filter-predict-one-step} until $t=4.5$.
The predictive variance for each sample is displayed as a $\pm 2$ standard deviation fill for each sample. For reference, the Lyapunov time for the deterministic latent system with the ground truth parameters is shown as a red vertical bar. 
This indicates the time for which prediction should be possible for the chaotic system \cite{WOLF1985285, shaw1981strange}.
The latent predictions from the misspecified prior model rapidly diverge from the true latent state. In comparison, the posterior predictions successfully track the latent state up until the Lyapunov time, where predictions become uncertain and begin to diverge as expected.
It should be noted the prior model is misspecified in the sense that the prior mean produces a non-chaotic system, with most of the probability mass concentrated on non-chaotic systems.

\subsection{Korteweg–de Vries PDE}
As previously, the latent transition density
$p(\vu_n | \vu_{n - 1}, \mLambda)$ defines the evolution of the FEM
coefficients, as given by a \gls*{statFEM} discretisation of a stochastic \gls*{KdV} equation:
\begin{gather*}
  \partial_t u + \alpha u \partial_s u  + \beta \partial_s^3 u + \nu u = \dot{\xi}, \quad
  \dot{\xi} \sim \mathcal{GP}(0, \delta(t - t') \cdot k(s, s')),
\end{gather*}
where $u := u(s, t)$, $\xi := \xi(s, t)$, $s\in [0, 2]$, $t\in [0,1]$, and $u(s, t) = u(s + 2, t)$.
We set the coefficients
$\alpha = 1, \beta = 0.022^2$, and we investigate two cases, one with and one without drag ($\nu=0, \nu=1$). The initial condition is set to
$u_h^{\mathrm{true}}(s, 0) = \cos(\pi s)$ as in the classical work
of~\citet{zabusky1965interaction}. This regime is characterised by the steepening
of the initial condition and the generation of
solitons; nonlinear waves which have particle-like interactions~\citep{drazin1989solitons}.

Data is generated by simulating a
trajectory $\vu_{1:N}^{\mathrm{true}}$ using a \gls*{FEM} discretisation of the
deterministic \gls*{KdV} equation and we impose \gls*{FEM} solutions
$u_h^{\mathrm{true}}(s, n\Delta_t)$ on a 2D grid. We light up pixels below the solution, which mimics an experimental setup where the wave-profile is recorded side-on. Independent salt-and-pepper noise is used to corrupt the image frames. 
We set the decoder as $p_\theta(\vy_n | \vx_n) = \mathrm{Bern}(\mu_\theta(\vx_n))$ and the encoder as $q_\phi(\vx_n | \vy_n) = \NPDF(\mu_\phi(\vy_n), \sigma_\phi(\vy_n))$; see \ref{app:numerics-and-architectures} for details.

\subsection*{\gls*{KdV} Results}

The aim of this experiment is to learn the mapping from image frames to the latent observations, which we approximate with a neural network as described in Section~\ref{subsec:variational-inference}.  Additionally, we wish to estimate the posterior distribution over the unknown drag coefficient $\nu$, for both with drag ($\nu=1$) and without ($\nu=0$).  The prior distribution $p(\nu)$ (with $\mLambda = \left\{  \nu \right\}$) implicitly places a prior over a family of \gls*{KdV} PDEs with varying drag coefficients.  Since this physical coefficient is positively constrained, we reflect this in the choice of prior such that $\text{supp}(p(\nu)) = \mathbb{R}^{+}$. For $\nu=1$ we use a log-normal prior distribution, and log-normal variational approximation, which we denote by $p(\nu) = \mathcal{LN}(\mu_{0}, \sigma_{0}^2)$, $q_{\lambda}(\nu) = \mathcal{LN}(\mu_{\lambda}, \sigma_{\lambda}^2)$ respectively.  For the no-drag case ($\nu=0$), we use the exponential distribution for both prior and variational approximation, denoted by $p(\nu) = \text{Exp}(\lambda_0)$ and $q_{\lambda}(\nu) = \text{Exp}(\lambda)$ respectively. For both cases, the choice of prior/posterior pairs provide analytic KL-divergences (see Table~\ref{tab:variational}) for use in the \gls*{ELBO} computation. Following the Lorenz-63 example, the variational approximations are initialised at poorly specified priors to assess the model's ability to correct for this misspecification. \\

\begin{figure}[ht!]
  \centering
  \begin{subfigure}[t]{0.28\textwidth}
    \centering
    \includegraphics[width=\textwidth]{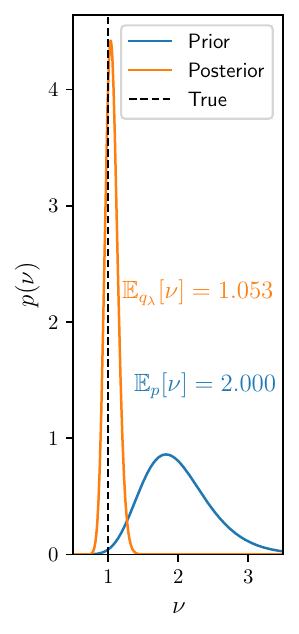}
    \caption{}
    \label{fig:kdv-estimation-lognormal}
  \end{subfigure}
  \begin{subfigure}[t]{0.70\textwidth}
      \centering
      \includegraphics[width=\textwidth]{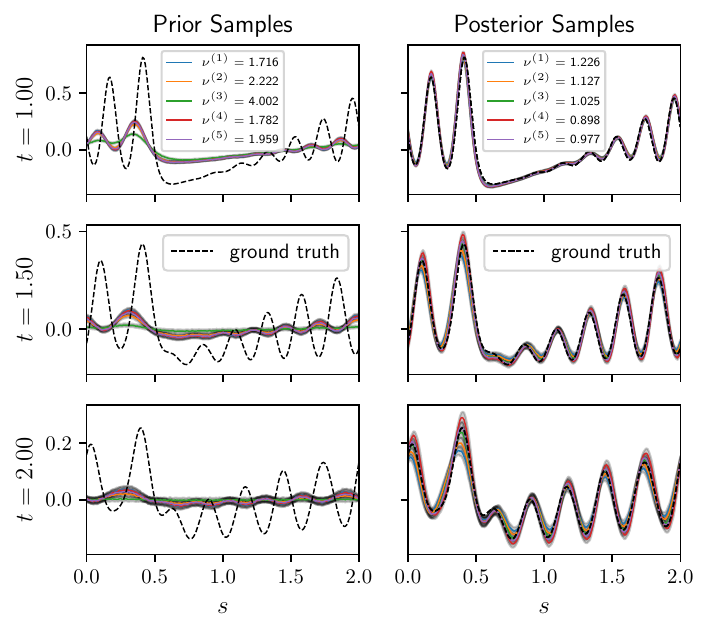}
      \caption{}
      \label{fig:kdv-lognormal-samples}
  \end{subfigure}
  \begin{subfigure}[t]{\textwidth}
      \centering
        \includegraphics[width=\textwidth]{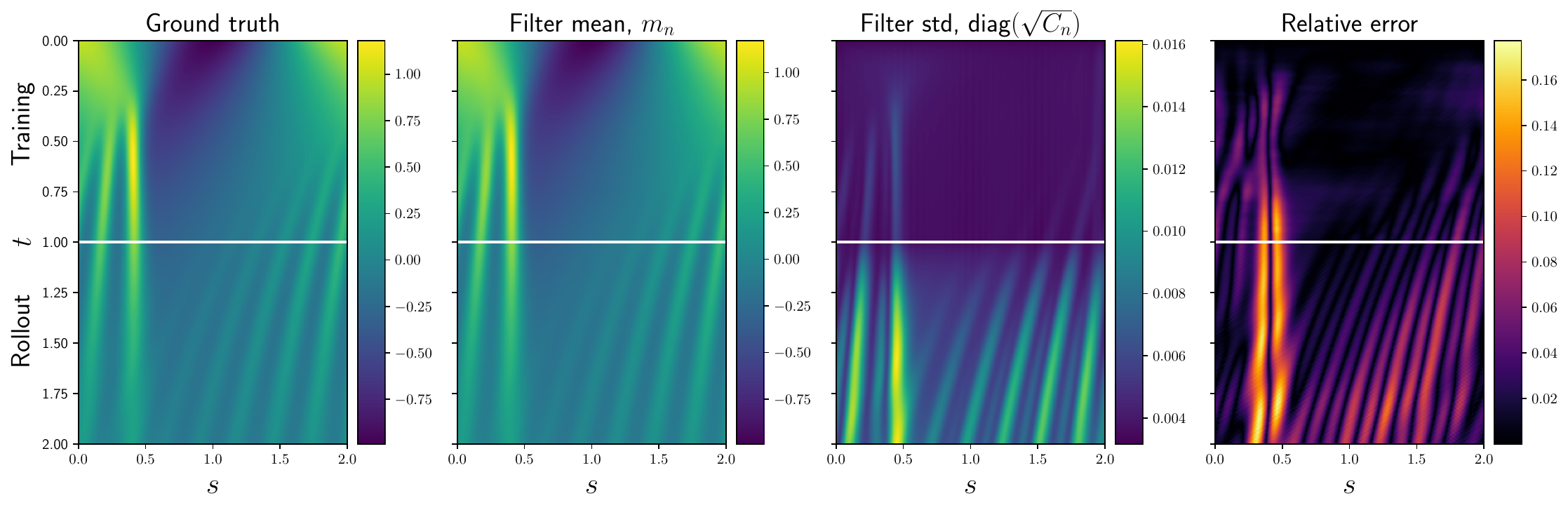}
      \caption{}
      \label{fig:kdv-inference-lognormal}
  \end{subfigure}
  \caption{Inference results for \gls*{KdV} with drag. (a) Comparison of prior and variational posterior for model parameter $\nu = 1$. (b) Latent filtering distribution shown for prior and posterior model for multiple samples. \textit{Prior model}: $\left\{ \phi, \theta \right\}$ trained using $\mLambda \sim  p(\nu)$. \textit{Posterior model}: $\left\{ \phi^{\star}, \theta^{\star}, \lambda^{\star} \right\}$ trained using $\mLambda \sim  q_{\lambda}(\nu)$. (c) Rollout prediction. The values of the latent process are shown in space (\textit{horizontal axis}) and time (\textit{vertical axis}). Rollout begins at $t=1.0$ (\textit{horizontal white line}). Relative error: $|\vu_n - \vu_n^{\text{true}}| / \max_{1\leq i \leq n_u} |u_{i, n}^{\text{true}}|$.
  }
  \label{fig:kdv-results-lognormal}
\end{figure}

\begin{figure}[ht!]
  \centering
  \begin{subfigure}[t]{0.28\textwidth}
    \centering
    \includegraphics[width=\textwidth]{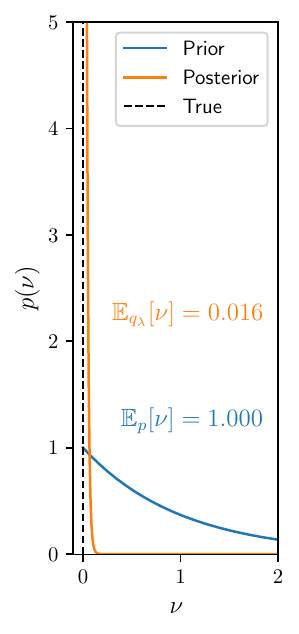}
    \caption{}
    \label{fig:kdv-estimation-exponential}
  \end{subfigure}
  \begin{subfigure}[t]{0.70\textwidth}
      \centering
      \includegraphics[width=\textwidth]{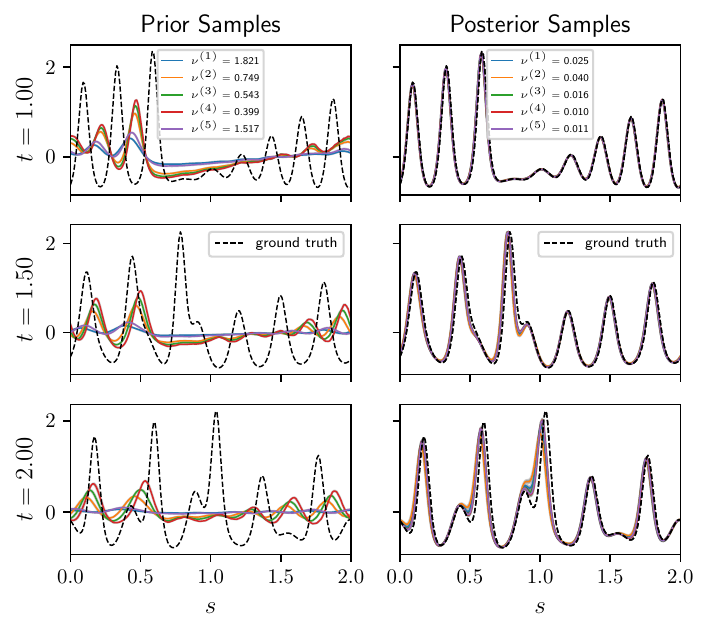}
      \caption{}
      \label{fig:kdv-exponential-samples}
      \end{subfigure}
  \begin{subfigure}[t]{\textwidth}
      \centering
      \includegraphics[width=\textwidth]{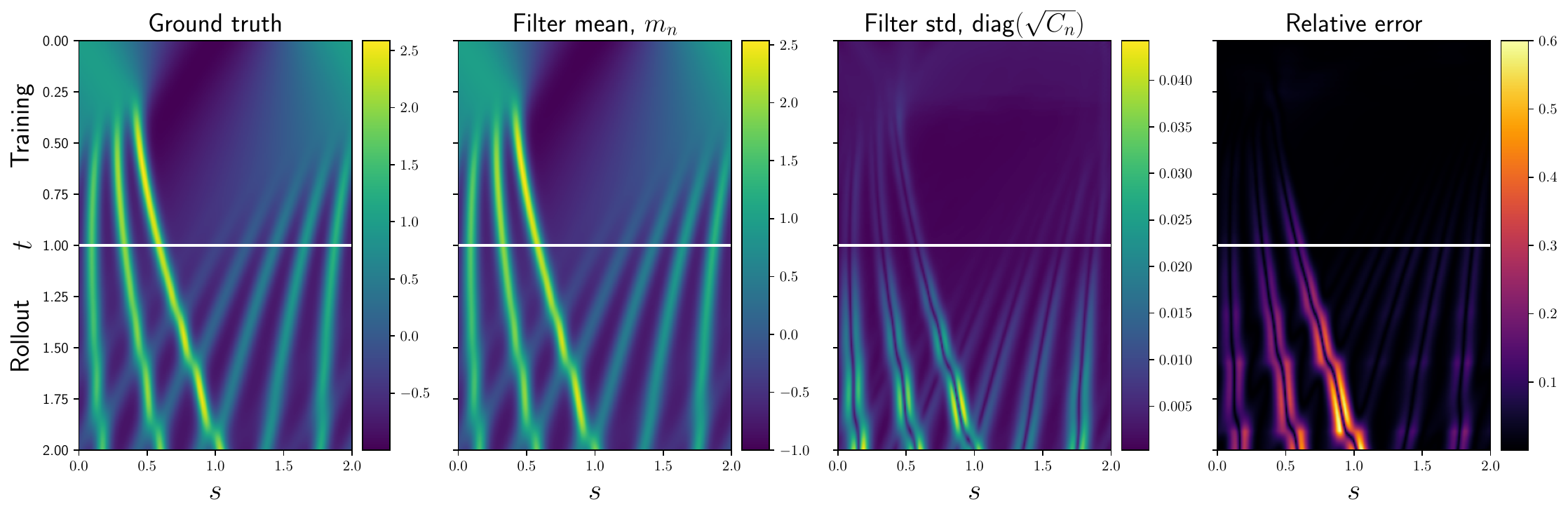}
      \caption{}
      \label{fig:kdv-inference-exponential}
      \end{subfigure}
  \caption{Inference results for no-drag \gls*{KdV}. (a) Comparison of prior and variational posterior for model parameter $\nu = 0$. (b) Latent filtering distribution shown for prior and posterior model for multiple samples. \textit{Prior model}: $\left\{ \phi, \theta \right\}$ trained using $\mLambda \sim  p(\nu)$. \textit{Posterior model}: $\left\{ \phi^{\star}, \theta^{\star}, \lambda^{\star} \right\}$ trained using $\mLambda \sim  q_{\lambda}(\nu)$. (c) Rollout prediction. The values of the latent process are shown in space (\textit{horizontal axis}) and time (\textit{vertical axis}). Rollout begins at $t=1.0$ (\textit{horizontal white line}). Relative error: $|\vu_n - \vu_n^{\text{true}}| / \max_{1\leq i \leq n_u} |u_{i, n}^{\text{true}}|$.
  }
  \label{fig:kdv-results-e}
\end{figure}

\noindent \textbf{\gls*{KdV} with drag} \quad For $\nu=1$, we set the prior $p(\nu) = \mathcal{LN}(2, 0.5^2)$ and show this alongside the posterior estimate in Fig.~\ref{fig:kdv-estimation-lognormal}. The posterior has contracted around the true value. 
Fig.~\ref{fig:kdv-lognormal-samples} shows the latent inference at specific time-points for multiple samples from both the prior and posterior, showing the improvement of the model when adapting the variational posterior jointly with the encoder. 
Marginalising over the parameter $\nu$ using samples from the posterior approximation $q_{\lambda^{\star}}(\nu)$ allow us to investigate the effect of uncertainty on the parameter estimate. These samples show a spread in mean estimates where the solitons are mixing at time $t=2.5$; the amplitude here is sensitive to changes in the drag coefficient, which is reflected in the variation of the filtering posterior samples. 
Fig.~\ref{fig:kdv-inference-lognormal} shows the resulting inference and rollout for the jointly trained encoder $\phi^{\star}$ and sample from the posterior $q_{\lambda^{\star}}(\nu)$. 
The relative error is computed by normalising the absolute error by the max absolute value of the true solution at each time point, which accounts for the reduction in amplitude of the solution through time so that the growth in relative error can be observed.
The latent mean is visualised alongside the ground truth, with the relative error. The relative error is low for the training time, and steadily increasing during rollout. Visualisation of the uncertainty is provided by the standard deviation of each latent dimension through time. The structure of the uncertainty is consistent with the relative error. 
We note that when initialised with a prior over $\nu$ centered on a much less dissipative system, the model struggles to adjust the posterior to a more dissipative system. We believe that since the pre-training on a less dissipative system produces latents that are more separated on the latent space, the autoencoder is able to still able to reconstruct the video frames due to the flexibility of the \gls*{NN} decoder. For these results, the system is initialised with a large drag coefficient, and early stopping used for regularisation. \\

\noindent \textbf{No-drag \gls*{KdV}} \quad For $\nu=0$, we set the prior $p(\nu) = \text{Exp}(1.0)$ and show this alongside the posterior estimate in Fig.~\ref{fig:kdv-estimation-exponential}. The posterior $q_{\lambda}(\nu) = \text{Exp}(62.2)$ 
places more density on the true value, $q_{\hat{\lambda}}(0)>p(0)$, and the resulting latent inference shows the comparison to the prior (see Fig.~\ref{fig:kdv-exponential-samples}). Whilst the posterior shows an improved fit to the true latents compared to the prior, there is still a clear mismatch. For $t=1$ when the rollout begins, the solitons of the latent samples appear out of phase with the true solution, which causes a mismatch when predicted forward in time. There is also a bias due to the parameter samples being non-zero, and hence predictions begin to diverge from the true solution which does not dissipate.
Fig.~\ref{fig:kdv-inference-exponential} shows the inference and rollout prediction. The predicted mean has the same structure as the true latents, and the covariance structure reflects the error in the mean.

\pagebreak
\section{Conclusions}
\label{sec:conclusion}

In this work, we introduced the concept of \acrfull*{UDA}, where unstructured data is used for state and parameter inference in a known physics model. We proposed a dynamical autoencoder approach to address the \gls*{UDA} problem, and demonstrated the model on three differential equation systems. The model can accurately embed unstructured data into pseudo-observations of the known physical system, and recover the true latent state of the system by filtering the pseudo-observations with the \acrfull*{ExKF}.

With the inclusion of prior knowledge on the physics, data-efficient learning is achieved. In particular, the extrapolation of the Lorenz-63 system to inferring unseen regions of the attractor based on $50$ velocity field observations, and the inversion of the \gls*{KdV} \gls*{PDE} with $100$ observed video frames. Accurate parameter estimation has also been demonstrated in both these examples, with the added benefit of parameter uncertainty quantification. We have shown the effect of maginalising out this uncertainty on the latent inference, which results in a mixture of Gaussians filtering distribution that is able to capture variation in the latent solution in areas that are sensitive to model parameters. Comparison of \gls*{PIDVAE} was made to a baseline model, the \gls*{KVAE}. This showed our model performing better in terms of Bayesian model evidence and \acrlong*{MSE} of reconstruction due to the added inductive bias introduced by the known physical system, compared to an arbitrary learned linear Gaussian state-space model. 

Our \gls*{PIDVAE} model has demonstrated that state estimation and inversion are possible in the case of \gls*{UDA}, when carefully embedding knowledge about the underlying dynamical system. Further work could extend this methodology to include \textit{direct} measurements, as well as the \textit{indirect}, unstructured observations in order to augment the inference when direct measurements are sparse, or expensive.

\section*{Acknowledgements}
We thank Andrew Wang for support with comparative experiments. A. Glyn-Davies was supported by Splunk Inc. [G106483] PhD scholarship funding. \"{O}. D. A. was partly supported by the Lloyd’s Register Foundation Data Centric Engineering Programme and EPSRC Programme Grant [EP/R034710/1] (CoSInES). C. Duffin and M. Girolami were supported by EPSRC grant [EP/T000414/1]. M. Girolami was supported by a Royal Academy of Engineering Research Chair, and EPSRC grants [EP/W005816/1], [EP/V056441/1], [EP/V056522/1], [EP/R018413/2], [EP/R034710/1], and [EP/R004889/1].

\bibliographystyle{elsarticle-num-names} 
\bibliography{references}

\begin{thebibliography}{89}
\expandafter\ifx\csname natexlab\endcsname\relax\def\natexlab#1{#1}\fi
\providecommand{\url}[1]{\texttt{#1}}
\providecommand{\href}[2]{#2}
\providecommand{\path}[1]{#1}
\providecommand{\DOIprefix}{doi:}
\providecommand{\ArXivprefix}{arXiv:}
\providecommand{\URLprefix}{URL: }
\providecommand{\Pubmedprefix}{pmid:}
\providecommand{\doi}[1]{\href{http://dx.doi.org/#1}{\path{#1}}}
\providecommand{\Pubmed}[1]{\href{pmid:#1}{\path{#1}}}
\providecommand{\bibinfo}[2]{#2}
\ifx\xfnm\relax \def\xfnm[#1]{\unskip,\space#1}\fi
\bibitem[{Judd and Smith(2004)}]{judd2004indistinguishable}
\bibinfo{author}{K.~Judd}, \bibinfo{author}{L.~Smith},
\newblock \bibinfo{title}{Indistinguishable states {{II}}. {{The}} imperfect
  model scenario},
\newblock \bibinfo{journal}{Physica D: Nonlinear Phenomena}
  \bibinfo{volume}{196} (\bibinfo{year}{2004}) \bibinfo{pages}{224--242}.
\bibitem[{Box(1979)}]{box1979robustness}
\bibinfo{author}{G.~E.~P. Box},
\newblock \bibinfo{title}{Robustness in the {{Strategy}} of {{Scientific Model
  Building}}},
\newblock in: \bibinfo{editor}{R.~L. Launer}, \bibinfo{editor}{G.~N. Wilkinson}
  (Eds.), \bibinfo{booktitle}{Robustness in {{Statistics}}},
  \bibinfo{publisher}{{Academic Press}}, \bibinfo{year}{1979}, pp.
  \bibinfo{pages}{201--236}.
\bibitem[{Kalman(1960)}]{kalman1960new}
\bibinfo{author}{R.~E. Kalman},
\newblock \bibinfo{title}{A {{New Approach}} to {{Linear Filtering}} and
  {{Prediction Problems}}},
\newblock \bibinfo{journal}{Journal of Basic Engineering} \bibinfo{volume}{82}
  (\bibinfo{year}{1960}) \bibinfo{pages}{35--45}.
\bibitem[{Anderson and Moore(1979)}]{anderson1979optimal}
\bibinfo{author}{B.~D. Anderson}, \bibinfo{author}{J.~B. Moore},
  \bibinfo{title}{Optimal filtering}, \bibinfo{publisher}{Englewood Cliffs,
  N.J. Prentice Hall}, \bibinfo{year}{1979}.
\bibitem[{Tarantola(2005)}]{tarantola2005inverse}
\bibinfo{author}{A.~Tarantola}, \bibinfo{title}{Inverse {{Problem Theory}} and
  {{Methods}} for {{Model Parameter Estimation}}}, \bibinfo{publisher}{{Society
  for Industrial and Applied Mathematics}}, \bibinfo{year}{2005}.
\bibitem[{Stuart(2010)}]{stuart2010inverse}
\bibinfo{author}{A.~M. Stuart},
\newblock \bibinfo{title}{Inverse problems: {{A Bayesian}} perspective},
\newblock \bibinfo{journal}{Acta Numerica} \bibinfo{volume}{19}
  (\bibinfo{year}{2010}) \bibinfo{pages}{451--559}.
\bibitem[{Law et~al.(2015)Law, Stuart, and Zygalakis}]{law2015data}
\bibinfo{author}{K.~Law}, \bibinfo{author}{A.~Stuart},
  \bibinfo{author}{K.~Zygalakis}, \bibinfo{title}{Data {{Assimilation}}: {{A
  Mathematical Introduction}}}, volume~\bibinfo{volume}{62},
  \bibinfo{publisher}{{Springer}}, \bibinfo{address}{{Cham, Switerland}},
  \bibinfo{year}{2015}.
\bibitem[{Reich and Cotter(2015)}]{reich2015probabilistic}
\bibinfo{author}{S.~Reich}, \bibinfo{author}{C.~Cotter},
  \bibinfo{title}{Probabilistic forecasting and Bayesian data assimilation},
  \bibinfo{publisher}{Cambridge University Press}, \bibinfo{year}{2015}.
\bibitem[{Kantas et~al.(2015)Kantas, Doucet, Singh, Maciejowski, and
  Chopin}]{kantas2015particle}
\bibinfo{author}{N.~Kantas}, \bibinfo{author}{A.~Doucet},
  \bibinfo{author}{S.~S. Singh}, \bibinfo{author}{J.~Maciejowski},
  \bibinfo{author}{N.~Chopin},
\newblock \bibinfo{title}{On particle methods for parameter estimation in
  state-space models},
\newblock \bibinfo{journal}{Statistical science} \bibinfo{volume}{30}
  (\bibinfo{year}{2015}) \bibinfo{pages}{328--351}.
\bibitem[{Storvik(2002)}]{storvik2002particle}
\bibinfo{author}{G.~Storvik},
\newblock \bibinfo{title}{Particle filters for state-space models with the
  presence of unknown static parameters},
\newblock \bibinfo{journal}{IEEE Transactions on signal Processing}
  \bibinfo{volume}{50} (\bibinfo{year}{2002}) \bibinfo{pages}{281--289}.
\bibitem[{Bocquet and Sakov(2013)}]{bocquet2013joint}
\bibinfo{author}{M.~Bocquet}, \bibinfo{author}{P.~Sakov},
\newblock \bibinfo{title}{Joint state and parameter estimation with an
  iterative ensemble kalman smoother},
\newblock \bibinfo{journal}{Nonlinear Processes in Geophysics}
  \bibinfo{volume}{20} (\bibinfo{year}{2013}) \bibinfo{pages}{803--818}.
\bibitem[{Ditlevsen and Samson(2014)}]{ditlevsen2014estimation}
\bibinfo{author}{S.~Ditlevsen}, \bibinfo{author}{A.~Samson},
\newblock \bibinfo{title}{Estimation in the partially observed stochastic
  morris--lecar neuronal model with particle filter and stochastic
  approximation methods},
\newblock \bibinfo{journal}{The annals of applied statistics}
  \bibinfo{volume}{8} (\bibinfo{year}{2014}) \bibinfo{pages}{674--702}.
\bibitem[{Kravaris et~al.(2013)Kravaris, Hahn, and Chu}]{kravaris2013advances}
\bibinfo{author}{C.~Kravaris}, \bibinfo{author}{J.~Hahn},
  \bibinfo{author}{Y.~Chu},
\newblock \bibinfo{title}{Advances and selected recent developments in state
  and parameter estimation},
\newblock \bibinfo{journal}{Computers \& chemical engineering}
  \bibinfo{volume}{51} (\bibinfo{year}{2013}) \bibinfo{pages}{111--123}.
\bibitem[{Dochain(2003)}]{dochain2003state}
\bibinfo{author}{D.~Dochain},
\newblock \bibinfo{title}{State and parameter estimation in chemical and
  biochemical processes: a tutorial},
\newblock \bibinfo{journal}{Journal of process control} \bibinfo{volume}{13}
  (\bibinfo{year}{2003}) \bibinfo{pages}{801--818}.
\bibitem[{Moradkhani et~al.(2005)Moradkhani, Sorooshian, Gupta, and
  Houser}]{moradkhani2005dual}
\bibinfo{author}{H.~Moradkhani}, \bibinfo{author}{S.~Sorooshian},
  \bibinfo{author}{H.~V. Gupta}, \bibinfo{author}{P.~R. Houser},
\newblock \bibinfo{title}{Dual state--parameter estimation of hydrological
  models using ensemble kalman filter},
\newblock \bibinfo{journal}{Advances in water resources} \bibinfo{volume}{28}
  (\bibinfo{year}{2005}) \bibinfo{pages}{135--147}.
\bibitem[{Blei et~al.(2017)Blei, Kucukelbir, and
  McAuliffe}]{blei2017variational}
\bibinfo{author}{D.~M. Blei}, \bibinfo{author}{A.~Kucukelbir},
  \bibinfo{author}{J.~D. McAuliffe},
\newblock \bibinfo{title}{Variational {{Inference}}: {{A Review}} for
  {{Statisticians}}},
\newblock \bibinfo{journal}{Journal of the American Statistical Association}
  \bibinfo{volume}{112} (\bibinfo{year}{2017}) \bibinfo{pages}{859--877}.
\bibitem[{Kingma and Welling(2014)}]{kingma2014autoencoding}
\bibinfo{author}{D.~P. Kingma}, \bibinfo{author}{M.~Welling},
  \bibinfo{title}{Auto-{{Encoding Variational Bayes}}}, \bibinfo{year}{2014}.
  \href{http://arxiv.org/abs/1312.6114}{{\tt arXiv:1312.6114}}.
\bibitem[{Fraccaro et~al.(2017)Fraccaro, Kamronn, Paquet, and
  Winther}]{fraccaro2017disentangled}
\bibinfo{author}{M.~Fraccaro}, \bibinfo{author}{S.~Kamronn},
  \bibinfo{author}{U.~Paquet}, \bibinfo{author}{O.~Winther},
\newblock \bibinfo{title}{A {{Disentangled Recognition}} and {{Nonlinear
  Dynamics Model}} for {{Unsupervised Learning}}},
\newblock in: \bibinfo{booktitle}{Advances in {{Neural Information Processing
  Systems}}}, volume~\bibinfo{volume}{30}, \bibinfo{publisher}{{Curran
  Associates, Inc.}}, \bibinfo{year}{2017}.
\bibitem[{Girin et~al.(2021)Girin, Leglaive, Bie, Diard, Hueber, and
  Alameda-Pineda}]{girin2021dynamical}
\bibinfo{author}{L.~Girin}, \bibinfo{author}{S.~Leglaive},
  \bibinfo{author}{X.~Bie}, \bibinfo{author}{J.~Diard},
  \bibinfo{author}{T.~Hueber}, \bibinfo{author}{X.~Alameda-Pineda},
\newblock \bibinfo{title}{Dynamical variational autoencoders: A comprehensive
  review},
\newblock \bibinfo{journal}{Foundations and Trends in Machine Learning}
  \bibinfo{volume}{15} (\bibinfo{year}{2021}) \bibinfo{pages}{1--175}.
\bibitem[{Pearce(2020)}]{pearce2020gaussian}
\bibinfo{author}{M.~Pearce},
\newblock \bibinfo{title}{The gaussian process prior vae for interpretable
  latent dynamics from pixels},
\newblock in: \bibinfo{booktitle}{Symposium on advances in approximate bayesian
  inference}, \bibinfo{year}{2020}, pp. \bibinfo{pages}{1--12}.
  \bibinfo{note}{Tex.organization: PMLR}.
\bibitem[{Jazbec et~al.(2021)Jazbec, Ashman, Fortuin, Pearce, Mandt, and
  R{\"a}tsch}]{jazbec2021scalable}
\bibinfo{author}{M.~Jazbec}, \bibinfo{author}{M.~Ashman},
  \bibinfo{author}{V.~Fortuin}, \bibinfo{author}{M.~Pearce},
  \bibinfo{author}{S.~Mandt}, \bibinfo{author}{G.~R{\"a}tsch},
\newblock \bibinfo{title}{Scalable {{Gaussian Process Variational
  Autoencoders}}},
\newblock in: \bibinfo{booktitle}{Proceedings of {{The}} 24th {{International
  Conference}} on {{Artificial Intelligence}} and {{Statistics}}},
  \bibinfo{publisher}{{PMLR}}, \bibinfo{year}{2021}, pp.
  \bibinfo{pages}{3511--3519}.
\bibitem[{Fortuin et~al.(2020)Fortuin, Baranchuk, R{\"a}tsch, and
  Mandt}]{fortuin2020gp}
\bibinfo{author}{V.~Fortuin}, \bibinfo{author}{D.~Baranchuk},
  \bibinfo{author}{G.~R{\"a}tsch}, \bibinfo{author}{S.~Mandt},
\newblock \bibinfo{title}{Gp-vae: Deep probabilistic time series imputation},
\newblock in: \bibinfo{booktitle}{International conference on artificial
  intelligence and statistics}, \bibinfo{organization}{PMLR},
  \bibinfo{year}{2020}, pp. \bibinfo{pages}{1651--1661}.
\bibitem[{Zhu et~al.(2022)Zhu, Rodas, and Li}]{zhu2022markovian}
\bibinfo{author}{H.~Zhu}, \bibinfo{author}{C.~B. Rodas},
  \bibinfo{author}{Y.~Li}, \bibinfo{title}{Markovian {{Gaussian Process
  Variational Autoencoders}}}, \bibinfo{year}{2022}.
  \href{http://arxiv.org/abs/2207.05543}{{\tt arXiv:2207.05543}}.
\bibitem[{Hartikainen and Sarkka(2010)}]{hartikainen2010kalman}
\bibinfo{author}{J.~Hartikainen}, \bibinfo{author}{S.~Sarkka},
\newblock \bibinfo{title}{Kalman filtering and smoothing solutions to temporal
  {{Gaussian}} process regression models},
\newblock in: \bibinfo{booktitle}{2010 {{IEEE International Workshop}} on
  {{Machine Learning}} for {{Signal Processing}}}, \bibinfo{publisher}{{IEEE}},
  \bibinfo{address}{{Kittila, Finland}}, \bibinfo{year}{2010}, pp.
  \bibinfo{pages}{379--384}.
\bibitem[{Yildiz et~al.(2019)Yildiz, Heinonen, and
  Lahdesmaki}]{yildiz2019ode2vae}
\bibinfo{author}{C.~Yildiz}, \bibinfo{author}{M.~Heinonen},
  \bibinfo{author}{H.~Lahdesmaki},
\newblock \bibinfo{title}{{{ODE2VAE}}: {{Deep}} generative second order
  {{ODEs}} with {{Bayesian}} neural networks},
\newblock in: \bibinfo{booktitle}{Advances in {{Neural Information Processing
  Systems}}}, volume~\bibinfo{volume}{32}, \bibinfo{publisher}{{Curran
  Associates, Inc.}}, \bibinfo{year}{2019}.
\bibitem[{Chen et~al.(2018)Chen, Rubanova, Bettencourt, and
  Duvenaud}]{chen2018neural}
\bibinfo{author}{R.~T.~Q. Chen}, \bibinfo{author}{Y.~Rubanova},
  \bibinfo{author}{J.~Bettencourt}, \bibinfo{author}{D.~K. Duvenaud},
\newblock \bibinfo{title}{Neural {{Ordinary Differential Equations}}},
\newblock in: \bibinfo{booktitle}{Advances in {{Neural Information Processing
  Systems}}}, volume~\bibinfo{volume}{31}, \bibinfo{publisher}{{Curran
  Associates, Inc.}}, \bibinfo{year}{2018}.
\bibitem[{Bayer and Osendorfer(2014)}]{bayer2014learning}
\bibinfo{author}{J.~Bayer}, \bibinfo{author}{C.~Osendorfer},
\newblock \bibinfo{title}{Learning stochastic recurrent networks},
\newblock \bibinfo{journal}{arXiv preprint arXiv:1411.7610}
  (\bibinfo{year}{2014}).
\bibitem[{Krishnan et~al.(2015)Krishnan, Shalit, and Sontag}]{krishnan2015deep}
\bibinfo{author}{R.~G. Krishnan}, \bibinfo{author}{U.~Shalit},
  \bibinfo{author}{D.~Sontag},
\newblock \bibinfo{title}{Deep kalman filters},
\newblock \bibinfo{journal}{arXiv preprint arXiv:1511.05121}
  (\bibinfo{year}{2015}).
\bibitem[{Karl et~al.(2017)Karl, Soelch, Bayer, and van~der
  Smagt}]{karl2017deep}
\bibinfo{author}{M.~Karl}, \bibinfo{author}{M.~Soelch},
  \bibinfo{author}{J.~Bayer}, \bibinfo{author}{P.~van~der Smagt},
\newblock \bibinfo{title}{Deep variational bayes filters: Unsupervised learning
  of state space models from raw data},
\newblock in: \bibinfo{booktitle}{International Conference on Learning
  Representations}, \bibinfo{year}{2017}.
\bibitem[{Wu et~al.(2021)Wu, Nair, {Martin-Martin}, {Fei-Fei}, and
  Finn}]{wu2021greedy}
\bibinfo{author}{B.~Wu}, \bibinfo{author}{S.~Nair},
  \bibinfo{author}{R.~{Martin-Martin}}, \bibinfo{author}{L.~{Fei-Fei}},
  \bibinfo{author}{C.~Finn},
\newblock \bibinfo{title}{Greedy {{Hierarchical Variational Autoencoders}} for
  {{Large-Scale Video Prediction}}},
\newblock in: \bibinfo{booktitle}{2021 {{IEEE}}/{{CVF Conference}} on
  {{Computer Vision}} and {{Pattern Recognition}} ({{CVPR}})},
  \bibinfo{publisher}{{IEEE}}, \bibinfo{address}{{Nashville, TN, USA}},
  \bibinfo{year}{2021}, pp. \bibinfo{pages}{2318--2328}.
\bibitem[{Franceschi et~al.(2020)Franceschi, Delasalles, Chen, Lamprier, and
  Gallinari}]{franceschi2020stochastic}
\bibinfo{author}{J.-Y. Franceschi}, \bibinfo{author}{E.~Delasalles},
  \bibinfo{author}{M.~Chen}, \bibinfo{author}{S.~Lamprier},
  \bibinfo{author}{P.~Gallinari},
\newblock \bibinfo{title}{Stochastic {{Latent Residual Video Prediction}}},
\newblock in: \bibinfo{booktitle}{Proceedings of the 37th {{International
  Conference}} on {{Machine Learning}}}, \bibinfo{publisher}{{PMLR}},
  \bibinfo{year}{2020}, pp. \bibinfo{pages}{3233--3246}.
\bibitem[{Babaeizadeh et~al.(2022)Babaeizadeh, Finn, Erhan, Campbell, and
  Levine}]{babaeizadeh2022stochastic}
\bibinfo{author}{M.~Babaeizadeh}, \bibinfo{author}{C.~Finn},
  \bibinfo{author}{D.~Erhan}, \bibinfo{author}{R.~H. Campbell},
  \bibinfo{author}{S.~Levine},
\newblock \bibinfo{title}{Stochastic {{Variational Video Prediction}}},
\newblock in: \bibinfo{booktitle}{International {{Conference}} on {{Learning
  Representations}}}, \bibinfo{year}{2022}.
\bibitem[{Chung et~al.(2015)Chung, Kastner, Dinh, Goel, Courville, and
  Bengio}]{chung2015recurrent}
\bibinfo{author}{J.~Chung}, \bibinfo{author}{K.~Kastner},
  \bibinfo{author}{L.~Dinh}, \bibinfo{author}{K.~Goel}, \bibinfo{author}{A.~C.
  Courville}, \bibinfo{author}{Y.~Bengio},
\newblock \bibinfo{title}{A recurrent latent variable model for sequential
  data},
\newblock \bibinfo{journal}{Advances in neural information processing systems}
  \bibinfo{volume}{28} (\bibinfo{year}{2015}).
\bibitem[{Brunton et~al.(2016)Brunton, Proctor, and
  Kutz}]{brunton2016discovering}
\bibinfo{author}{S.~L. Brunton}, \bibinfo{author}{J.~L. Proctor},
  \bibinfo{author}{J.~N. Kutz},
\newblock \bibinfo{title}{Discovering governing equations from data by sparse
  identification of nonlinear dynamical systems},
\newblock \bibinfo{journal}{Proceedings of the national academy of sciences}
  \bibinfo{volume}{113} (\bibinfo{year}{2016}) \bibinfo{pages}{3932--3937}.
\bibitem[{Champion et~al.(2019)Champion, Lusch, Kutz, and
  Brunton}]{champion2019data}
\bibinfo{author}{K.~Champion}, \bibinfo{author}{B.~Lusch},
  \bibinfo{author}{J.~N. Kutz}, \bibinfo{author}{S.~L. Brunton},
\newblock \bibinfo{title}{Data-driven discovery of coordinates and governing
  equations},
\newblock \bibinfo{journal}{Proceedings of the National Academy of Sciences}
  \bibinfo{volume}{116} (\bibinfo{year}{2019}) \bibinfo{pages}{22445--22451}.
\bibitem[{Lopez and Atzberger(2021)}]{lopez2021variational}
\bibinfo{author}{R.~Lopez}, \bibinfo{author}{P.~J. Atzberger},
  \bibinfo{title}{Variational {{Autoencoders}} for {{Learning Nonlinear
  Dynamics}} of {{Physical Systems}}}, \bibinfo{year}{2021}.
  \href{http://arxiv.org/abs/2012.03448}{{\tt arXiv:2012.03448}}.
\bibitem[{Lusch et~al.(2018)Lusch, Kutz, and Brunton}]{lusch2018deep}
\bibinfo{author}{B.~Lusch}, \bibinfo{author}{J.~N. Kutz},
  \bibinfo{author}{S.~L. Brunton},
\newblock \bibinfo{title}{Deep learning for universal linear embeddings of
  nonlinear dynamics},
\newblock \bibinfo{journal}{Nature communications} \bibinfo{volume}{9}
  (\bibinfo{year}{2018}) \bibinfo{pages}{1--10}.
\bibitem[{Otto and Rowley(2019)}]{otto2019linearly}
\bibinfo{author}{S.~E. Otto}, \bibinfo{author}{C.~W. Rowley},
\newblock \bibinfo{title}{Linearly {{Recurrent Autoencoder Networks}} for
  {{Learning Dynamics}}},
\newblock \bibinfo{journal}{SIAM Journal on Applied Dynamical Systems}
  \bibinfo{volume}{18} (\bibinfo{year}{2019}) \bibinfo{pages}{558--593}.
\bibitem[{Gin et~al.(2021)Gin, Lusch, Brunton, and Kutz}]{gin2021deep}
\bibinfo{author}{C.~Gin}, \bibinfo{author}{B.~Lusch}, \bibinfo{author}{S.~L.
  Brunton}, \bibinfo{author}{J.~N. Kutz},
\newblock \bibinfo{title}{Deep learning models for global coordinate
  transformations that linearise {{PDEs}}},
\newblock \bibinfo{journal}{European Journal of Applied Mathematics}
  \bibinfo{volume}{32} (\bibinfo{year}{2021}) \bibinfo{pages}{515--539}.
\bibitem[{Morton et~al.(2018)Morton, Jameson, Kochenderfer, and
  Witherden}]{morton2018deep}
\bibinfo{author}{J.~Morton}, \bibinfo{author}{A.~Jameson},
  \bibinfo{author}{M.~J. Kochenderfer}, \bibinfo{author}{F.~Witherden},
\newblock \bibinfo{title}{Deep {{Dynamical Modeling}} and {{Control}} of
  {{Unsteady Fluid Flows}}},
\newblock in: \bibinfo{booktitle}{Advances in {{Neural Information Processing
  Systems}}}, volume~\bibinfo{volume}{31}, \bibinfo{publisher}{{Curran
  Associates, Inc.}}, \bibinfo{year}{2018}.
\bibitem[{Takeishi et~al.(2017)Takeishi, Kawahara, and
  Yairi}]{takeishi2017learning}
\bibinfo{author}{N.~Takeishi}, \bibinfo{author}{Y.~Kawahara},
  \bibinfo{author}{T.~Yairi},
\newblock \bibinfo{title}{Learning {{Koopman Invariant Subspaces}} for
  {{Dynamic Mode Decomposition}}},
\newblock in: \bibinfo{booktitle}{Advances in {{Neural Information Processing
  Systems}}}, volume~\bibinfo{volume}{30}, \bibinfo{publisher}{{Curran
  Associates, Inc.}}, \bibinfo{year}{2017}.
\bibitem[{Erichson et~al.(2019)Erichson, Muehlebach, and
  Mahoney}]{erichson2019physicsinformed}
\bibinfo{author}{N.~B. Erichson}, \bibinfo{author}{M.~Muehlebach},
  \bibinfo{author}{M.~W. Mahoney}, \bibinfo{title}{Physics-informed
  {{Autoencoders}} for {{Lyapunov-stable Fluid Flow Prediction}}},
  \bibinfo{year}{2019}. \href{http://arxiv.org/abs/1905.10866}{{\tt
  arXiv:1905.10866}}.
\bibitem[{Hern{\'a}ndez et~al.(2018)Hern{\'a}ndez, {Wayment-Steele}, Sultan,
  Husic, and Pande}]{hernandez2018variational}
\bibinfo{author}{C.~X. Hern{\'a}ndez}, \bibinfo{author}{H.~K.
  {Wayment-Steele}}, \bibinfo{author}{M.~M. Sultan}, \bibinfo{author}{B.~E.
  Husic}, \bibinfo{author}{V.~S. Pande},
\newblock \bibinfo{title}{Variational encoding of complex dynamics},
\newblock \bibinfo{journal}{Physical Review E} \bibinfo{volume}{97}
  (\bibinfo{year}{2018}) \bibinfo{pages}{062412}.
\bibitem[{Lu et~al.(2020)Lu, Kim, and Solja{\v c}i{\'c}}]{lu2020extracting}
\bibinfo{author}{P.~Y. Lu}, \bibinfo{author}{S.~Kim},
  \bibinfo{author}{M.~Solja{\v c}i{\'c}},
\newblock \bibinfo{title}{Extracting {{Interpretable Physical Parameters}} from
  {{Spatiotemporal Systems Using Unsupervised Learning}}},
\newblock \bibinfo{journal}{Physical Review X} \bibinfo{volume}{10}
  (\bibinfo{year}{2020}) \bibinfo{pages}{031056}.
\bibitem[{Yin et~al.(2021)Yin, Le~Guen, Dona, {de B{\'e}zenac}, Ayed, Thome,
  and Gallinari}]{yin2021augmenting}
\bibinfo{author}{Y.~Yin}, \bibinfo{author}{V.~Le~Guen},
  \bibinfo{author}{J.~Dona}, \bibinfo{author}{E.~{de B{\'e}zenac}},
  \bibinfo{author}{I.~Ayed}, \bibinfo{author}{N.~Thome},
  \bibinfo{author}{P.~Gallinari},
\newblock \bibinfo{title}{Augmenting physical models with deep networks for
  complex dynamics forecasting},
\newblock \bibinfo{journal}{Journal of Statistical Mechanics: Theory and
  Experiment} \bibinfo{volume}{2021} (\bibinfo{year}{2021})
  \bibinfo{pages}{124012}.
\bibitem[{Long et~al.(2018)Long, Lu, Ma, and Dong}]{long2018pdeneta}
\bibinfo{author}{Z.~Long}, \bibinfo{author}{Y.~Lu}, \bibinfo{author}{X.~Ma},
  \bibinfo{author}{B.~Dong},
\newblock \bibinfo{title}{{{PDE-Net}}: {{Learning PDEs}} from {{Data}}},
\newblock in: \bibinfo{booktitle}{Proceedings of the 35th {{International
  Conference}} on {{Machine Learning}}}, \bibinfo{publisher}{{PMLR}},
  \bibinfo{year}{2018}, pp. \bibinfo{pages}{3208--3216}.
\bibitem[{de~B{\'e}zenac et~al.(2019)de~B{\'e}zenac, Pajot, and
  Gallinari}]{bezenac2019deep}
\bibinfo{author}{E.~de~B{\'e}zenac}, \bibinfo{author}{A.~Pajot},
  \bibinfo{author}{P.~Gallinari},
\newblock \bibinfo{title}{Deep learning for physical processes: Incorporating
  prior scientific knowledge},
\newblock \bibinfo{journal}{Journal of Statistical Mechanics: Theory and
  Experiment} \bibinfo{volume}{2019} (\bibinfo{year}{2019})
  \bibinfo{pages}{124009}.
\bibitem[{Shin and Choi(2023)}]{SHIN2023112183}
\bibinfo{author}{H.~Shin}, \bibinfo{author}{M.~Choi},
\newblock \bibinfo{title}{Physics-informed variational inference for
  uncertainty quantification of stochastic differential equations},
\newblock \bibinfo{journal}{Journal of Computational Physics}
  \bibinfo{volume}{487} (\bibinfo{year}{2023}) \bibinfo{pages}{112183}.
\bibitem[{Goh et~al.(2022)Goh, Sheriffdeen, Wittmer, and
  Bui-Thanh}]{goh2022solving}
\bibinfo{author}{H.~Goh}, \bibinfo{author}{S.~Sheriffdeen},
  \bibinfo{author}{J.~Wittmer}, \bibinfo{author}{T.~Bui-Thanh},
\newblock \bibinfo{title}{Solving bayesian inverse problems via variational
  autoencoders},
\newblock volume \bibinfo{volume}{145} of \textit{\bibinfo{series}{Proceedings
  of Machine Learning Research}}, \bibinfo{publisher}{PMLR},
  \bibinfo{year}{2022}, pp. \bibinfo{pages}{386--425}.
\bibitem[{Raissi et~al.(2019)Raissi, Perdikaris, and
  Karniadakis}]{RAISSI2019686}
\bibinfo{author}{M.~Raissi}, \bibinfo{author}{P.~Perdikaris},
  \bibinfo{author}{G.~Karniadakis},
\newblock \bibinfo{title}{Physics-informed neural networks: A deep learning
  framework for solving forward and inverse problems involving nonlinear
  partial differential equations},
\newblock \bibinfo{journal}{Journal of Computational Physics}
  \bibinfo{volume}{378} (\bibinfo{year}{2019}) \bibinfo{pages}{686--707}.
\bibitem[{Wandel et~al.(2022)Wandel, Weinmann, Neidlin, and
  Klein}]{Wandel20228529}
\bibinfo{author}{N.~Wandel}, \bibinfo{author}{M.~Weinmann},
  \bibinfo{author}{M.~Neidlin}, \bibinfo{author}{R.~Klein},
\newblock \bibinfo{title}{Spline-pinn: Approaching pdes without data using
  fast, physics-informed hermite-spline cnns},
\newblock volume~\bibinfo{volume}{36}, \bibinfo{year}{2022}, p.
  \bibinfo{pages}{8529 – 8538}.
\bibitem[{Zhu et~al.(2019)Zhu, Zabaras, Koutsourelakis, and
  Perdikaris}]{ZHU201956}
\bibinfo{author}{Y.~Zhu}, \bibinfo{author}{N.~Zabaras}, \bibinfo{author}{P.-S.
  Koutsourelakis}, \bibinfo{author}{P.~Perdikaris},
\newblock \bibinfo{title}{Physics-constrained deep learning for
  high-dimensional surrogate modeling and uncertainty quantification without
  labeled data},
\newblock \bibinfo{journal}{Journal of Computational Physics}
  \bibinfo{volume}{394} (\bibinfo{year}{2019}) \bibinfo{pages}{56--81}.
\bibitem[{Yang et~al.(2021)Yang, Meng, and Karniadakis}]{YANG2021109913}
\bibinfo{author}{L.~Yang}, \bibinfo{author}{X.~Meng}, \bibinfo{author}{G.~E.
  Karniadakis},
\newblock \bibinfo{title}{B-pinns: Bayesian physics-informed neural networks
  for forward and inverse pde problems with noisy data},
\newblock \bibinfo{journal}{Journal of Computational Physics}
  \bibinfo{volume}{425} (\bibinfo{year}{2021}) \bibinfo{pages}{109913}.
\bibitem[{Meng et~al.(2022)Meng, Seo, Cao, Griesemer, and
  Liu}]{meng2022physics}
\bibinfo{author}{C.~Meng}, \bibinfo{author}{S.~Seo}, \bibinfo{author}{D.~Cao},
  \bibinfo{author}{S.~Griesemer}, \bibinfo{author}{Y.~Liu},
\newblock \bibinfo{title}{When physics meets machine learning: A survey of
  physics-informed machine learning},
\newblock \bibinfo{journal}{arXiv preprint arXiv:2203.16797}
  (\bibinfo{year}{2022}).
\bibitem[{Karniadakis et~al.(2021)Karniadakis, Kevrekidis, Lu, Perdikaris,
  Wang, and Yang}]{karniadakis2021physics}
\bibinfo{author}{G.~E. Karniadakis}, \bibinfo{author}{I.~G. Kevrekidis},
  \bibinfo{author}{L.~Lu}, \bibinfo{author}{P.~Perdikaris},
  \bibinfo{author}{S.~Wang}, \bibinfo{author}{L.~Yang},
\newblock \bibinfo{title}{Physics-informed machine learning},
\newblock \bibinfo{journal}{Nature Reviews Physics} \bibinfo{volume}{3}
  (\bibinfo{year}{2021}) \bibinfo{pages}{422--440}.
\bibitem[{Geneva and Zabaras(2020)}]{geneva2020modeling}
\bibinfo{author}{N.~Geneva}, \bibinfo{author}{N.~Zabaras},
\newblock \bibinfo{title}{Modeling the dynamics of pde systems with
  physics-constrained deep auto-regressive networks},
\newblock \bibinfo{journal}{Journal of Computational Physics}
  \bibinfo{volume}{403} (\bibinfo{year}{2020}) \bibinfo{pages}{109056}.
\bibitem[{Cicirello(2024)}]{cicirello2024physics}
\bibinfo{author}{A.~Cicirello},
\newblock \bibinfo{title}{Physics-enhanced machine learning: a position paper
  for dynamical systems investigations},
\newblock \bibinfo{journal}{arXiv preprint arXiv:2405.05987}
  (\bibinfo{year}{2024}).
\bibitem[{Lathourakis and Cicirello(2024)}]{lathourakis2024physics}
\bibinfo{author}{C.~Lathourakis}, \bibinfo{author}{A.~Cicirello},
\newblock \bibinfo{title}{Physics enhanced sparse identification of dynamical
  systems with discontinuous nonlinearities},
\newblock \bibinfo{journal}{Nonlinear Dynamics}  (\bibinfo{year}{2024})
  \bibinfo{pages}{1--28}.
\bibitem[{Kovachki et~al.(2023)Kovachki, Li, Liu, Azizzadenesheli,
  Bhattacharya, Stuart, and Anandkumar}]{kovachki2023neural}
\bibinfo{author}{N.~Kovachki}, \bibinfo{author}{Z.~Li},
  \bibinfo{author}{B.~Liu}, \bibinfo{author}{K.~Azizzadenesheli},
  \bibinfo{author}{K.~Bhattacharya}, \bibinfo{author}{A.~Stuart},
  \bibinfo{author}{A.~Anandkumar},
\newblock \bibinfo{title}{Neural operator: Learning maps between function
  spaces with applications to pdes},
\newblock \bibinfo{journal}{Journal of Machine Learning Research}
  \bibinfo{volume}{24} (\bibinfo{year}{2023}) \bibinfo{pages}{1--97}.
\bibitem[{Li et~al.(2020)Li, Kovachki, Azizzadenesheli, Liu, Bhattacharya,
  Stuart, and Anandkumar}]{li2020fourier}
\bibinfo{author}{Z.~Li}, \bibinfo{author}{N.~Kovachki},
  \bibinfo{author}{K.~Azizzadenesheli}, \bibinfo{author}{B.~Liu},
  \bibinfo{author}{K.~Bhattacharya}, \bibinfo{author}{A.~Stuart},
  \bibinfo{author}{A.~Anandkumar},
\newblock \bibinfo{title}{Fourier neural operator for parametric partial
  differential equations},
\newblock \bibinfo{journal}{arXiv preprint arXiv:2010.08895}
  (\bibinfo{year}{2020}).
\bibitem[{Li et~al.(2021)Li, Zheng, Kovachki, Jin, Chen, Liu, Azizzadenesheli,
  and Anandkumar}]{li2021physics}
\bibinfo{author}{Z.~Li}, \bibinfo{author}{H.~Zheng},
  \bibinfo{author}{N.~Kovachki}, \bibinfo{author}{D.~Jin},
  \bibinfo{author}{H.~Chen}, \bibinfo{author}{B.~Liu},
  \bibinfo{author}{K.~Azizzadenesheli}, \bibinfo{author}{A.~Anandkumar},
\newblock \bibinfo{title}{Physics-informed neural operator for learning partial
  differential equations},
\newblock \bibinfo{journal}{arXiv preprint arXiv:2111.03794}
  (\bibinfo{year}{2021}).
\bibitem[{Vadeboncoeur et~al.(2022)Vadeboncoeur, Akyildiz, Kazlauskaite,
  Girolami, and Cirak}]{vadeboncoeur2022deep}
\bibinfo{author}{A.~Vadeboncoeur}, \bibinfo{author}{{\"O}.~D. Akyildiz},
  \bibinfo{author}{I.~Kazlauskaite}, \bibinfo{author}{M.~Girolami},
  \bibinfo{author}{F.~Cirak},
\newblock \bibinfo{title}{Deep probabilistic models for forward and inverse
  problems in parametric pdes},
\newblock \bibinfo{journal}{arXiv preprint arXiv:2208.04856}
  (\bibinfo{year}{2022}).
\bibitem[{Vadeboncoeur et~al.(2023)Vadeboncoeur, Kazlauskaite, Papandreou,
  Cirak, Girolami, and Akyildiz}]{vadeboncoeur2023random}
\bibinfo{author}{A.~Vadeboncoeur}, \bibinfo{author}{I.~Kazlauskaite},
  \bibinfo{author}{Y.~Papandreou}, \bibinfo{author}{F.~Cirak},
  \bibinfo{author}{M.~Girolami}, \bibinfo{author}{O.~D. Akyildiz},
\newblock \bibinfo{title}{Random grid neural processes for parametric partial
  differential equations},
\newblock in: \bibinfo{booktitle}{International Conference on Machine
  Learning}, \bibinfo{organization}{PMLR}, \bibinfo{year}{2023}, pp.
  \bibinfo{pages}{34759--34778}.
\bibitem[{Girolami et~al.(2021)Girolami, Febrianto, Yin, and
  Cirak}]{girolami2021statistical}
\bibinfo{author}{M.~Girolami}, \bibinfo{author}{E.~Febrianto},
  \bibinfo{author}{G.~Yin}, \bibinfo{author}{F.~Cirak},
\newblock \bibinfo{title}{The statistical finite element method ({{statFEM}})
  for coherent synthesis of observation data and model predictions},
\newblock \bibinfo{journal}{Computer Methods in Applied Mechanics and
  Engineering} \bibinfo{volume}{375} (\bibinfo{year}{2021})
  \bibinfo{pages}{113533}.
\bibitem[{Duffin et~al.(2021)Duffin, Cripps, Stemler, and
  Girolami}]{duffin2021statistical}
\bibinfo{author}{C.~Duffin}, \bibinfo{author}{E.~Cripps},
  \bibinfo{author}{T.~Stemler}, \bibinfo{author}{M.~Girolami},
\newblock \bibinfo{title}{Statistical finite elements for misspecified models},
\newblock \bibinfo{journal}{Proceedings of the National Academy of Sciences}
  \bibinfo{volume}{118} (\bibinfo{year}{2021}).
\bibitem[{Duffin et~al.(2022)Duffin, Cripps, Stemler, and
  Girolami}]{duffin2022lowrank}
\bibinfo{author}{C.~Duffin}, \bibinfo{author}{E.~Cripps},
  \bibinfo{author}{T.~Stemler}, \bibinfo{author}{M.~Girolami},
\newblock \bibinfo{title}{Low-rank statistical finite elements for scalable
  model-data synthesis},
\newblock \bibinfo{journal}{Journal of Computational Physics}
  \bibinfo{volume}{463} (\bibinfo{year}{2022}).
\bibitem[{Akyildiz et~al.(2022)Akyildiz, Duffin, Sabanis, and
  Girolami}]{akyildiz2022statistical}
\bibinfo{author}{{\"O}.~D. Akyildiz}, \bibinfo{author}{C.~Duffin},
  \bibinfo{author}{S.~Sabanis}, \bibinfo{author}{M.~Girolami},
\newblock \bibinfo{title}{Statistical {F}inite {E}lements via {L}angevin
  {D}ynamics},
\newblock \bibinfo{journal}{SIAM/ASA Journal of Uncertainty Quantification}
  (\bibinfo{year}{2022}).
\bibitem[{Williams and Rasmussen(2006)}]{williams2006gaussian}
\bibinfo{author}{C.~K. Williams}, \bibinfo{author}{C.~E. Rasmussen},
  \bibinfo{title}{Gaussian Processes for Machine Learning},
  volume~\bibinfo{volume}{2}, \bibinfo{publisher}{{MIT press Cambridge, MA}},
  \bibinfo{year}{2006}.
\bibitem[{Lalchand and Rasmussen(2020)}]{lalchand2020approximate}
\bibinfo{author}{V.~Lalchand}, \bibinfo{author}{C.~E. Rasmussen},
\newblock \bibinfo{title}{Approximate inference for fully bayesian gaussian
  process regression},
\newblock in: \bibinfo{booktitle}{Symposium on Advances in Approximate Bayesian
  Inference}, \bibinfo{organization}{PMLR}, \bibinfo{year}{2020}, pp.
  \bibinfo{pages}{1--12}.
\bibitem[{Drazin and Johnson(1989)}]{drazin1989solitons}
\bibinfo{author}{P.~G. Drazin}, \bibinfo{author}{R.~S. Johnson},
  \bibinfo{title}{Solitons: {{An Introduction}}}, Cambridge {{Texts}} in
  {{Applied Mathematics}}, \bibinfo{edition}{second} ed.,
  \bibinfo{publisher}{{Cambridge University Press}},
  \bibinfo{address}{{Cambridge}}, \bibinfo{year}{1989}.
\bibitem[{Brenner and Scott(2008)}]{brenner2008mathematical}
\bibinfo{author}{S.~C. Brenner}, \bibinfo{author}{L.~R. Scott},
  \bibinfo{title}{The {{Mathematical Theory}} of {{Finite Element Methods}}},
  volume~\bibinfo{volume}{15} of \textit{\bibinfo{series}{Texts in {{Applied
  Mathematics}}}}, \bibinfo{publisher}{{Springer New York}},
  \bibinfo{address}{{New York, NY}}, \bibinfo{year}{2008}.
\bibitem[{Thom{\'e}e(2006)}]{thomeeGalerkinFiniteElement2006}
\bibinfo{author}{V.~Thom{\'e}e}, \bibinfo{title}{Galerkin Finite Element
  Methods for Parabolic Problems}, number \bibinfo{number}{v. 25} in
  \bibinfo{series}{Springer Series in Computational Mathematics},
  \bibinfo{edition}{2nd ed} ed., \bibinfo{publisher}{{Springer}},
  \bibinfo{address}{{Berlin ; New York}}, \bibinfo{year}{2006}.
\bibitem[{Jazwinski(1970)}]{jazwinski1970stochastic}
\bibinfo{author}{A.~H. Jazwinski}, \bibinfo{title}{Stochastic Processes and
  Filtering Theory}, \bibinfo{publisher}{Academic Press}, \bibinfo{year}{1970}.
\bibitem[{Chen et~al.(2022)Chen, Sanz-Alonso, and
  Willett}]{chen2022autodifferentiable}
\bibinfo{author}{Y.~Chen}, \bibinfo{author}{D.~Sanz-Alonso},
  \bibinfo{author}{R.~Willett},
\newblock \bibinfo{title}{Autodifferentiable ensemble kalman filters},
\newblock \bibinfo{journal}{SIAM Journal on Mathematics of Data Science}
  \bibinfo{volume}{4} (\bibinfo{year}{2022}) \bibinfo{pages}{801--833}.
\bibitem[{Corenflos et~al.(2021)Corenflos, Thornton, Deligiannidis, and
  Doucet}]{corenflos2021differentiable}
\bibinfo{author}{A.~Corenflos}, \bibinfo{author}{J.~Thornton},
  \bibinfo{author}{G.~Deligiannidis}, \bibinfo{author}{A.~Doucet},
\newblock \bibinfo{title}{Differentiable particle filtering via
  entropy-regularized optimal transport},
\newblock in: \bibinfo{booktitle}{International Conference on Machine
  Learning}, \bibinfo{organization}{PMLR}, \bibinfo{year}{2021}, pp.
  \bibinfo{pages}{2100--2111}.
\bibitem[{Kingma et~al.(2019)Kingma, Welling et~al.}]{kingma2019introduction}
\bibinfo{author}{D.~P. Kingma}, \bibinfo{author}{M.~Welling}, et~al.,
\newblock \bibinfo{title}{An introduction to variational autoencoders},
\newblock \bibinfo{journal}{Foundations and Trends{\textregistered} in Machine
  Learning} \bibinfo{volume}{12} (\bibinfo{year}{2019})
  \bibinfo{pages}{307--392}.
\bibitem[{Kingma and Ba(2017)}]{kingma2017adam}
\bibinfo{author}{D.~P. Kingma}, \bibinfo{author}{J.~Ba}, \bibinfo{title}{Adam:
  {{A Method}} for {{Stochastic Optimization}}}, \bibinfo{year}{2017}.
  \href{http://arxiv.org/abs/1412.6980}{{\tt arXiv:1412.6980}}.
\bibitem[{Boncelet(2009)}]{boncelet2009image}
\bibinfo{author}{C.~Boncelet},
\newblock \bibinfo{title}{Chapter 7 - image noise models},
\newblock in: \bibinfo{booktitle}{The Essential Guide to Image Processing},
  \bibinfo{publisher}{Academic Press}, \bibinfo{address}{Boston},
  \bibinfo{year}{2009}, pp. \bibinfo{pages}{143--167}. \URLprefix
  \url{https://www.sciencedirect.com/science/article/pii/B978012374457900007X}.
  \DOIprefix\doi{https://doi.org/10.1016/B978-0-12-374457-9.00007-X}.
\bibitem[{Kloeden and Platen(1992)}]{kloedenpeter1992numerical}
\bibinfo{author}{P.~Kloeden}, \bibinfo{author}{E.~Platen},
  \bibinfo{title}{Numerical {{Solution}} of {{Stochastic Differential
  Equations}}}, Applications of {{Mathematics}}, \bibinfo{publisher}{{Springer
  Berlin Heidelberg}}, \bibinfo{year}{1992}.
\bibitem[{{\O}ksendal(2003)}]{oksendal2003stochastic}
\bibinfo{author}{B.~{\O}ksendal},
\newblock \bibinfo{title}{Stochastic differential equations},
\newblock in: \bibinfo{booktitle}{Stochastic Differential Equations},
  \bibinfo{publisher}{{Springer}}, \bibinfo{year}{2003}, pp.
  \bibinfo{pages}{65--84}.
\bibitem[{Lorenz(1963)}]{lorenz1963deterministic}
\bibinfo{author}{E.~N. Lorenz},
\newblock \bibinfo{title}{Deterministic nonperiodic flow},
\newblock \bibinfo{journal}{Journal of atmospheric sciences}
  \bibinfo{volume}{20} (\bibinfo{year}{1963}) \bibinfo{pages}{130--141}.
\bibitem[{Wolf et~al.(1985)Wolf, Swift, Swinney, and Vastano}]{WOLF1985285}
\bibinfo{author}{A.~Wolf}, \bibinfo{author}{J.~B. Swift},
  \bibinfo{author}{H.~L. Swinney}, \bibinfo{author}{J.~A. Vastano},
\newblock \bibinfo{title}{Determining lyapunov exponents from a time series},
\newblock \bibinfo{journal}{Physica D: Nonlinear Phenomena}
  \bibinfo{volume}{16} (\bibinfo{year}{1985}) \bibinfo{pages}{285--317}.
\bibitem[{Shaw(1981)}]{shaw1981strange}
\bibinfo{author}{R.~Shaw},
\newblock \bibinfo{title}{Strange attractors, chaotic behavior, and information
  flow},
\newblock \bibinfo{journal}{Zeitschrift f{\"u}r Naturforschung A}
  \bibinfo{volume}{36} (\bibinfo{year}{1981}) \bibinfo{pages}{80--112}.
\bibitem[{Zabusky and Kruskal(1965)}]{zabusky1965interaction}
\bibinfo{author}{N.~J. Zabusky}, \bibinfo{author}{M.~D. Kruskal},
\newblock \bibinfo{title}{Interaction of "{{Solitons}}" in a {{Collisionless
  Plasma}} and the {{Recurrence}} of {{Initial States}}},
\newblock \bibinfo{journal}{Physical Review Letters} \bibinfo{volume}{15}
  (\bibinfo{year}{1965}) \bibinfo{pages}{240--243}.
\bibitem[{Debussche and Printems(1999)}]{debussche1999numerical}
\bibinfo{author}{A.~Debussche}, \bibinfo{author}{J.~Printems},
\newblock \bibinfo{title}{Numerical simulation of the stochastic
  {{Korteweg}}\textendash de {{Vries}} equation},
\newblock \bibinfo{journal}{Physica D: Nonlinear Phenomena}
  \bibinfo{volume}{134} (\bibinfo{year}{1999}) \bibinfo{pages}{200--226}.
\bibitem[{Schiesser(1991)}]{schiesser1991numerical}
\bibinfo{author}{W.~E. Schiesser}, \bibinfo{title}{The {{Numerical Method}} of
  {{Lines}}: {{Integration}} of {{Partial Differential Equations}}},
  \bibinfo{publisher}{{Academic Press}}, \bibinfo{year}{1991}.
\bibitem[{Evans(2010)}]{evans2010partial}
\bibinfo{author}{L.~Evans}, \bibinfo{title}{Partial {{Differential
  Equations}}}, volume~\bibinfo{volume}{19} of
  \textit{\bibinfo{series}{Graduate {{Studies}} in {{Mathematics}}}},
  \bibinfo{edition}{second} ed., \bibinfo{publisher}{{American Mathematical
  Society}}, \bibinfo{address}{{Providence, Rhode Island}},
  \bibinfo{year}{2010}.
\bibitem[{Evensen(2003)}]{evensen2003ensemble}
\bibinfo{author}{G.~Evensen},
\newblock \bibinfo{title}{The {{Ensemble Kalman Filter}}: Theoretical
  formulation and practical implementation},
\newblock \bibinfo{journal}{Ocean Dynamics} \bibinfo{volume}{53}
  (\bibinfo{year}{2003}) \bibinfo{pages}{343--367}.
\bibitem[{Doucet et~al.(2000)Doucet, Godsill, and
  Andrieu}]{doucet2000sequential}
\bibinfo{author}{A.~Doucet}, \bibinfo{author}{S.~Godsill},
  \bibinfo{author}{C.~Andrieu},
\newblock \bibinfo{title}{On sequential {{Monte Carlo}} sampling methods for
  {{Bayesian}} filtering},
\newblock \bibinfo{journal}{Statistics and Computing} \bibinfo{volume}{10}
  (\bibinfo{year}{2000}) \bibinfo{pages}{197--208}.

\end{thebibliography}

\newpage

\appendix

\section{Numerical Details and Network Architectures}
\label{app:numerics-and-architectures}

\noindent \textbf{Lorenz-63 experiment.}
\begin{itemize}
  \item Time-series length: $N = 300$. Training length: $50$, Test length: $250$
  \item Input: $n_y = 200$. 
  \item Pseudo-observations: $n_x = 1$.
  \item Latent $n_u = 3$.
  \item Decoder: $p_{\vw}(\vy_n | x_n) = \NPDF(\vw x_n, \eta^2 \mI)$, $\eta = 0.1$.
  \item Encoder: $q_{\vw}(x_n|\vy_n) = \NPDF((\vw^\top \vw)^{-1} \vw^{\top} \vy_n, \eta^2 (\vw^\top \vw)^{-1})$ 
  \item Latent initial condition: $\vu_0 = \left[-3.7277, -3.8239, 21.1507\right]^\top$
  \item Latent noise processes: $\mL = \diag(1.0^2)$, $\mR = \diag(0.4^2)$.
  \item Latent discretisation: Euler-Maruyama, $dt = 0.001$.
  \item Model parameter prior: $p(\sigma) = \mathcal{N}(30, 5^2)$.
  \item Optimiser: Adam, learning rate = $10^{-2}$.
  \item Training Epochs: 2000.
\end{itemize}
To generate the data, we simulate the Lorenz SDE with $dt=0.001$ and take pseudo-observations $x_n$ every $40$ time-steps of the latent system, for a total of $N=300$ with $\Delta_t=0.04$. Velocity measurements are taken in the $s_1$ and $s_2$ directions over a regular $10 \times 10$ grid on the domain $s_1, s_2 \in [-4, 4]$, via the streamfunction $\psi(s_1, s_2, t)$. These measurements are flattened to the data vector $\vy_n \in \mathbb{R}^{n_y}$, $n_y = 200$. Parameters for data generation are $\mLambda = \left\{ \sigma=10, r=28, b=8/3 \right\}$.\\

\noindent \textbf{Advection equation experiment.}
\begin{itemize}
\item Time-series length: $N = 200$.
\item Input: $n_y = 784$. 
\item Pseudo-observations: $n_x = 64$.
\item Latent $n_u = 64$.
\item Decoder: $p_{\theta}(\vy_n | \vx_n) = \text{Bern}(\mu_{\theta}(\vx_n))$
\item $\mu_{\theta}(\cdot)$: MLP, two fully connected hidden layers with dimension 128
\item Encoder: $q_\phi(\vx_n | \vy_n) = \NPDF(\mu_\phi(\vy_n), \sigma_\phi(\vy_n))$
\item $\mu_{\phi}(\cdot), \log\sigma_{\phi}(\cdot)$: MLP, two fully connected hidden layers with dimension 128
\item Neural Network activations: LeakyReLU, negative slope $= 0.01$
\item Latent initial condition: $u(s, 0) = \exp(-2(s-2.5)^2), s\in[0,5]$.
\item Latent noise processes: $\rho = 0.1$, $\ell = 0.1$, $\mR = \diag(0.1^2)$.
\item Latent discretisation: \gls{FEM}, $C^0([0, 1])$ polynomial trial/test functions,
  Crank-Nicolson time discretisation, $dt = 0.02$.
\item Optimiser: Adam, learning rate = $0.001$.
\item Salt-and-pepper noise, probability of pixel information loss: $0.02$
\end{itemize}
To generate the data, we evaluate the analytical solution at every $dt=0.02$, taking pseudo-observations every 10 time-steps for $\Delta_t = 0.2$ and $N=200$. Latent dimensions are $n_u = 64$ and $n_x = 64$, with each image $28 \times 28$ pixels. These images are then flattened to vectors $\vy_n \in \left[0,1\right]^{n_y}$, with $n_y = 784$.\\

\noindent \textbf{\gls*{KdV} equation experiment.}
\begin{itemize}
\item Time-series length: $N = 100$.
\item Input: $n_y = 1792$.
\item Pseudo-observations: $n_x = 40$.
\item Latent $n_u = 600$.
\item Decoder: $p_{\theta}(\vy_n | \vx_n) = \text{Bern}(S(f_{\theta}(\vx_n)))$
\item $f_{\theta}(\cdot)$: MLP, three fully connected hidden layers with dimension 128
\item Encoder: $q_\phi(\vx_n | \vy_n) = \NPDF(\mu_\phi(\vy_n), \sigma_\phi(\vy_n))$
\item $\mu_{\phi}(\cdot), \log \sigma_{\phi}(\cdot)$: MLP, three fully connected hidden layers with dimension 128
\item Neural Network activations: tanh
\item Fixed parameters: $\alpha = 1.0, \beta = 0.022^2$
\item Unknown parameter (with drag): $\nu = 1.0$.
\item Unknown parameter (no-drag): $\nu = 0$.
\item Latent initial condition: $u(s, 0) = \cos(\pi s), s\in [0,2]$.
\item Latent noise processes: $\rho = 0.01$, $\ell = 0.2$, $\mR = \diag(0.02^2)$.
\item Latent discretisation: Petrov-Galerkin approach of~\citet{debussche1999numerical}:
  $C^0([0, 2])$ polynomial trial functions, Crank-Nicolson time
  discretisation, $dt = 0.01$.
\item Parameter prior (with drag): $p(\mLambda) = \mathcal{LN}(2, 0.5^2)$.
\item Parameter prior (no-drag): $p(\mLambda) = \text{Exp}(1.0)$.
\item Optimiser: Adam, learning rate = $0.005$.
\item Salt-and-pepper noise, probability of pixel information loss: $0.02$
\end{itemize}
To generate the data we simulate the \gls*{KdV} equation with a high-fidelity in space and time to ensure accurate latent data, and avoid inverse crimes. For the data used in comparison, we subsample this simulation with $dt = 0.01$, observing every timestep for $\Delta_t = 0.01$, and we take $N = 100$ observations $\vy_n$, with $\vy_n \in \left[0,1\right]^{n_y}$. Each $\vy_n$ is a flattened image of dimension $n_y = 64 \times 28 = 1792$, and we encode to pseudo-observations $\vx_n$ of dimension $n_x = 40$. The latent state dimension is $n_u = 600$. \\

\noindent\textbf{Neural network decoding/encoding.} \quad Variational autoencoders use neural networks to parameterise the encoding distribution $q_{\phi}(\vx|\vy)$, and the probablisitic decoder $p_{\theta}(\vy|\vx)$. A common choice of variational distribution is the Gaussian, which uses a \gls*{NN} to parameterise the mean $\mu_{\phi}(\vy)$, where $\mu_{\phi} \colon \bR^{n_y} \to \bR^{n_x}$, and another to parametrise the diagonal of the covariance. Parameterising the log-variance allows for both positive and negative neural network output --- the values are exponentiated to calculate the positively constrained variances.

\begin{align*}
  q_{\phi} (\vx|\vy) = \mathcal{N} \left( \mu_{\phi}(\vy), \text{diag}(\sigma_{\phi}(\vy)) \right)
\end{align*}

The decoder attempts to reconstruct the original data $\vy$ from the latents $\vx$. A \gls*{NN} decoder maps the latents to a reconstruction $\hat{\vy} = f_{\theta}(\vx)$, where $f_{\theta} \colon \bR^{n_x} \to \bR^{n_y}$ is the \gls*{NN} mapping parameterised by $\theta$. The probabilistic decoder depends on the form of the data. In this work, we use a Gaussian distribution for the velocity field data, and a Bernoulli distribution for the video data. When reconstructing video frames, an element-wise sigmoid function is used as the final layer of the neural network, in order to output grayscale pixel values, i.e. $S(y) = 1/(1+\exp (-y))$, where $S \colon \bR^{n_y} \to \left[0, 1\right]^{n_y}$. The two examples of unstructured data, real-valued velocity field data, and binary pixel data can be reconstructed as follows:\\
Real-valued data, $\vy \in \bR^{n_y}$, we reconstruct $\hat{\vy} \in \bR^{n_y}$, where $\hat{\vy} = f_{\theta}(\vx)$. Often in practice an independent noise model is used, with standard deviation $\sigma$, giving the density and log-density as follows: 
\begin{align*}
   p_{\theta} (\vy|\vx) &= \mathcal{N} \left(\vy ; \hat{\vy}, \sigma^2 \mI \right)\\
   \log p_{\theta} (\vy|\vx) &= - \frac{n_y}{2} \log(2 \pi \sigma^2) - \frac{1}{2\sigma^2} \left\Vert \vy - \hat{\vy} \right\Vert_2^2
\end{align*}

Binary pixel data, $\vy \in \left\{0,1\right\}^{n_y}$ we reconstruct $\hat{\vy} \in \left[0,1\right]^{n_y}$, where $\hat{\vy} = S\left(f_{\theta}(\vx)\right)$. The density and log-density are given as follows:
\begin{align*} 
    p_{\theta} (\vy|\vx) &= \text{Bern} \left(\vy; \hat{\vy} \right)\\
   \log p_{\theta} (\vy|\vx) &= \sum_{i=1}^{n_y} y_i \log (\hat{y}_i) + (1-y_i) \log(1-\hat{y}_i)
\end{align*}

\noindent\textbf{Linear decoding/encoding.} \quad Unlike the nonlinear \gls*{NN} decoders, if linear data generation is assumed from $\vx$ to $\vy$, we can invert this in order to encode our data. With a linear decoder of the form:
\begin{align*}
  p_{\mA}(\vy|\vx) = \mathcal{N}(\mA\vx, \eta^2 I).
\end{align*}
In this case, we use the ``inverted'' linear decoder given by:
\begin{align*}
  q_{\mA}(\vx|\vy) = \mathcal{N}((\mA^\top\mA)^{-1}\mA^\top\vy, \eta^2 (\mA^\top\mA)^{-1}).
\end{align*}
By selecting the encoder appropriately, the space of parameterised variational
distributions can be restricted to align with our beliefs about the data
generation process.

\section{Full Variational Framework}
\label{app:complete-variational-derivation}

We derive the approximate ELBO for joint estimation of model parameters $\mLambda$, and autoencoder parameters $\phi, \theta$. We start by writing the evidence
\begin{align*}
    p(\vy_{1:N}) = \int p(\vu_{1:N}, \vx_{1:N}, \mLambda, \vy_{1:N}) \md \vx_{1:N} \md \vu_{1:N} \md \mLambda.
\end{align*}
We maximise $\log p(\vy_{1:N})$ as
\begin{align*}
  \log p(\vy_{1:N}) &= \log \int p(\vu_{1:N}, 
  \vx_{1:N}, \mLambda, \vy_{1:N}) \md \vx_{1:N}d \vu_{1:N} \md \mLambda\\
                  &= \log \int \frac{p(\vu_{1:N}, \vx_{1:N}, \mLambda , \vy_{1:N})}{q(\vu_{1:N}, \vx_{1:N}, \mLambda | \vy_{1:N})} q(\vu_{1:N}, \vx_{1:N}, \mLambda | \vy_{1:N}) \md \vx_{1:N} \md \vu_{1:N} \md\mLambda\\
                  &\geq  \int \log \left[  \frac{p(\vu_{1:N}, \vx_{1:N}, \mLambda , \vy_{1:N})}{q(\vu_{1:N}, \vx_{1:N}, \mLambda | \vy_{1:N})}\right] q(
                 \vu_{1:N}, \vx_{1:N}, \mLambda | \vy_{1:N})  \md \vx_{1:N} \md \vu_{1:N} \md \mLambda\\
                    &= \text{ELBO},
\end{align*}
where the third line follows from the application of Jensen's inequality. Our
generative model determines the factorisation of the joint distribution, given in~\eqref{eq:factor-joint}:
\begin{align*}
  p(\vu_{1:N}, \vx_{1:N}, \mLambda , \vy_{1:N}) &= p(\vy_{1:N} | \vx_{1:N}, \vu_{1:N}, \mLambda) p(\vx_{1:N} | \vu_{1:N}, \mLambda) p(\vu_{1:N} | \mLambda)p(\mLambda)\\
  &= p_{\theta}(\vy_{1:N} | \vx_{1:N}) p(\vx_{1:N} | \vu_{1:N}, \mLambda)p(\vu_{1:N} | \mLambda)p(\mLambda).
\end{align*}
Next, we plug this factorised distribution into the ELBO and obtain
\begin{align*}
  \text{ELBO} = \int \log \left[  \frac{p_{\theta}(\vy_{1:N} | \vx_{1:N})
      p(\vx_{1:N} | \vu_{1:N}, \mLambda) p(\vu_{1:N} |
      \mLambda)p(\mLambda)}{q(\vu_{1:N}, \vx_{1:N}, \mLambda | \vy_{1:N})} \right] 
  \times q(\vu_{1:N}, \vx_{1:N}, \mLambda | \vy_{1:N}) \, \md \vx_{1:N} \md \vu_{1:N} \md \mLambda.
\end{align*}
The family of distributions which we use to approximate the posterior is
described below. We assume a factorisation based on the model into variational
encoding $q_{\phi}(\cdot)$, a full latent state posterior $q_{\nu}(\cdot)$, and the
variational approximation to the parameter posterior $q_{\lambda}(\cdot)$:
\begin{align*}
  q(\vu_{1:N}, \vx_{1:N}, \mLambda | 
  \vy_{1:N}) &=q(\vu_{1:N} | \vx_{1:N}, \mLambda) q_\lambda(\mLambda) q_{\phi}(\vx_{1:N} | \vy_{1:N}) \\
             &=q(\vu_{1:N} | \vx_{1:N}, \mLambda) q_\lambda(\mLambda)\prod_{n=1}^{N}q_{\phi}(\vx_n|\vy_n).
\end{align*}
The second line demonstrates the amortized structure of the autoencoder, where the same encoding parameters are shared across datapoints. We can then substitute this expression into our ELBO and obtain
\begin{align*}
  \text{ELBO} = \int \log \left[
  \frac{p_{\theta}(\vy_{1:N} | \vx_{1:N}) p(\vx_{1:N} | \vu_{1:N}, \mLambda) p(\vu_{1:N} | \mLambda) p(\mLambda)}{q(\vu_{1:N} | \vx_{1:N}, \mLambda) q_\lambda(\mLambda) q_{\phi}(\vx_{1:N} | \vy_{1:N})} \right] 
  \times q(\vu_{1:N} | \vx_{1:N}, \mLambda) q_\lambda(\mLambda) q_{\phi}(\vx_{1:N} | \vy_{1:N}) \md \vx_{1:N} \md \vu_{1:N} \md\mLambda.
\end{align*}
Assuming the variational posterior is the exact filtering posterior, i.e., $q(\vu_{1:N} | \vx_{1:N}, \mLambda) = p(\vu_{1:N} | \vx_{1:N}, \mLambda)$ then applying Bayes rule
\begin{align*}
    \frac{p(\vx_{1:N} | \vu_{1:N}, \mLambda) p(\vu_{1:N} | \mLambda)}{q(\vu_{1:N} | \vx_{1:N}, \mLambda) } = p(\vx_{1:N} | \mLambda)
\end{align*}
leads to a simplification of ELBO in terms of the marginal likelihood $p(\vx_{1:N} | \mLambda)$ of the state-space model. Substituting this expression leads to
\begin{align*}
  \text{ELBO} &= \int \log \left[
                    \frac{p_{\theta}(\vy_{1:N} | \vx_{1:N}) p(\vx_{1:N}|\mLambda)p(\mLambda)}{q_{\phi}( \vx_{1:N} | \vy_{1:N})q_\lambda(\mLambda)} \right] \\
              &\quad \quad \quad \quad \times q(\vu_{1:N} | \vx_{1:N}, \mLambda) q_\lambda(\mLambda)q_{\phi}(\vx_{1:N}|\vy_{1:N}) \md \vx_{1:N} \md \vu_{1:N} \md\mLambda \\
              &= \int \log \left[
                \frac{p_{\theta}(\vy_{1:N} | \vx_{1:N})}{q_{\phi}(\vx_{1:N}|\vy_{1:N})}
                \right]
                q_{\phi}(\vx_{1:N}|\vy_{1:N}) \md \vx_{1:N} \\ 
                & \quad \quad \quad \quad + \int \left[\log 
                p(\vx_{1:N}|\mLambda) + 
                \log   
                \frac{p(\mLambda)}{q_{\lambda}(\mLambda)}
                \right]
                q_{\lambda}(\mLambda) q_{\phi}(\vx_{1:N}|\vy_{1:N}) \md \vx_{1:N} \md\mLambda \\
              &= \mathbb{E}_{q_{\phi}} \left[   \log 
                \frac{p_{\theta}(\vy_{1:N} | \vx_{1:N})}{q_{\phi}(\vx_{1:N}|\vy_{1:N})}
                \right] +
                \mathbb{E}_{q_{\phi}} \left[
                \mathbb{E}_{q_{\lambda}}
                \left[\log 
                p(\vx_{1:N}|\mLambda)
                + 
                \log   
                \frac{p(\mLambda)}{q_\lambda(\mLambda)}
                \right]
                \right]\\
              &= \mathbb{E}_{q_{\phi}} \left[   \log 
                \frac{p_{\theta}(\vy_{1:N} | \vx_{1:N})}{q_{\phi}(\vx_{1:N}|\vy_{1:N})}
                \right] +
                \mathbb{E}_{q_{\phi}} \left[
                \mathbb{E}_{q_{\lambda}}
                \left[\log 
                p(\vx_{1:N}|\mLambda)
                \right]
                \right] - \kl{q_{\lambda}}{p}.
\end{align*}
Using a single MC samples $\vx_{1:N} \sim q_{\phi}(\cdot| \vy_{1:N})$, and $\mLambda\sim q_{\lambda}(\cdot)$ to approximate the
expectations, we can write the approximate \gls*{ELBO}, $\mathcal{L}(\phi, \theta, \lambda)$ as
\begin{align*}
  \mathcal{L}(\theta, \phi, \lambda) = \log p_{\theta}(\vy_{1:N} | \vx_{1:N}) - \log q_{\phi}(\vx_{1:N} | \vy_{1:N}) + 
  \log 
  p(\vx_{1:N}|\mLambda) - \kl{q_{\lambda}}{p}.
\end{align*}

See Algorithm~\ref{algo:phi-dvae} for pseudo-code, and Table~\ref{tab:variational} for distributions that provide analytic KL-divergence.


\section{Further Details on the Dynamic Model}
\label{app:complete-dynamical-model}

In this work we take the latent dynamical model to be a stochastic \gls*{ODE} or
\gls*{PDE}. For an ODE this follows from a standard
\gls*{SDE}~\citep{oksendal2003stochastic},
given by
\[
  \md \vu = f_\Lambda(\vu, t; \Lambda) \md t + \mL(t) \md \mW(t),
\]
where $\vu := \vu(t) \in \bR^{n_u}$, $t \in [0, T]$, $f_\Lambda \colon \bR^{n_u} \times [0,
T] \to \bR^{n_u}$, $\mL \colon [0, T] \to \bR^{n_u \times n_u}$. The noise process
$\mW(t)$ is a standard vector Brownian motion. The diffusion term $\mL(t)$ can be
used to describe any \textit{a priori} correlation in the error process
dimensions. As stated in the main text, this error process is taken to represent
possibly misspecified/unknown physics, which may have been omitted when
specifying the model. Discretisation with an explicit Euler-Maruyama scheme~\citep{kloedenpeter1992numerical}
gives,
\[
  \vu_n = \vu_{n - 1} + \Delta_t f_{n - 1}(\vu_{n - 1}; \Lambda) + \mL_{n - 1}
  \Delta \mW_{n - 1}, \quad \Delta \mW_{n - 1} \sim \NPDF(\bm{0}, \Delta_t \mI),
\]
where $\vu_n := \vu(n \Delta_t)$, $f_n(\cdot; \Lambda) = f_\Lambda(\cdot, n\Delta_t)$, and so on.
This gives a transition density
\[
  p(\vu_n | \vu_{n - 1}, \Lambda) = \NPDF(\vu_{n - 1} + \Delta_t f_{n - 1}(\vu_{n - 1}; \Lambda),
  \Delta_t \mL_{n - 1} \mL_{n - 1}^\top),
\]
defining a Markov model on the now discretised state vector $\vu_n$. To align
with the notation introduced in the main text, this gives:
  \begin{gather*}
    p(\vu_n | \vu_{n - 1}, \Lambda) = \NPDF(\cM(\vu_{n - 1}), \mQ), \\
    \cM(\vu_{n - 1}) := \vu_{n - 1} + \Delta_t f_{n - 1}(\vu_{n - 1}; \Lambda), \quad
    \mQ := \Delta_t \mL_{n - 1} \mL_{n - 1}^\top.
  \end{gather*}
Due to the structure of the \gls*{statFEM} discretisation, the
fully-discretised underlying model is of the same mathematical form as this \gls*{ODE} case. The
difference lies in the dynamics being defined from either a \gls*{PDE} or
\gls*{ODE} system. In common cases, a lower dimensional state vector, $\vu_n$,
typically results for the \gls*{ODE} case in comparison to the \gls*{PDE} case.
For the \gls*{PDE} case, entries of the state vector $\vu_n$ are coefficients of
the finite element basis functions.

For the PDE case, the derivation is similar, with an additional step
pre-time-discretisation to spatially discretise the system. This yields a
method-of-lines approach~\citep{schiesser1991numerical}. As in the main text, we consider a generic
nonlinear PDE system of the form
\begin{equation}
  \label{eq:PDE-formulation-appx}
  \partial_t u  + L_{\Lambda} u  + F_\Lambda(u) = f + \dot{\xi}, \quad
  \dot{\xi} \sim \mathcal{GP}(0, \delta(t-t') \cdot k(\vs,\vs')),
\end{equation}
where $u:=u(\vs, t)$, $\xi := \xi(\vs, t)$, $f:= f(\vs)$, $\vs \in \Omega \subset \bR^d$,
and $t \in [0,T]$. The operators $L_\Lambda$ and
$F_\Lambda(\cdot)$ are linear and nonlinear differential operators,
respectively. The process $\dot{\xi}$ is the derivative of a function-valued
Wiener process, whose increments are given by a Gaussian process with the
covariance kernel $k(\cdot, \cdot)$. In our examples, we use the
squared-exponential covariance function~\citep{williams2006gaussian}
\[
  k(\vs,\vs') = \rho^2 \exp\left(- \frac{\lVert \vs - \vs' \rVert_2^2}{2 \ell^2} \right).
\]
Hyperparameters $\{\rho, \ell\}$ are always assumed to be known, being set
\textit{a priori}. Further work investigating inference of these hyperparameters is of interest.

As stated in the main text we discretise the linear time-evolving PDE following
the \gls*{statFEM} as in \citet{duffin2021statistical}, for which we refer to
for the full details of this approach. In brief, we discretise spatially with finite elements 
(see, e.g.,~\citet{brenner2008mathematical,thomeeGalerkinFiniteElement2006}, for standard references) 
then temporally via finite differences. We first multiply
\eqref{eq:PDE-formulation-appx} with a
sufficiently smooth test function $v \in V$, where $V$ is an appropriate
function space (e.g. the $H_0^1(\Omega)$ Sobolev space~\citep{evans2010partial}) and
integrate over the domain $\Omega$ to give the weak form~\citep{brenner2008mathematical}
\[
  \langle \partial_t u, v \rangle + \cA_{\Lambda} (u, v) + \langle F_\Lambda(u),
  v \rangle = \langle f, v \rangle + \langle \dot{\xi}, v \rangle, \quad \forall v \in V.
\]
Recall that $\cA_\Lambda(\cdot, \cdot)$ is the bilinear form generated
from the linear operator $L_\Lambda$, and 
\[
\langle f, g \rangle = \int_\Omega f(\vs) g(\vs) \, \md \vs,
\]
the $L^2(\Omega)$ inner product.

Next we introduce a discrete approximation to the
domain, $\Omega_h \subseteq \Omega$, having vertices $\{ \vs_j \}_{j =
  1}^{n_h}$. This is parameterised by $h$ which indicates the degree of
mesh-refinement. We now introduce a finite-dimensional set of polynomial basis
functions $\{\phi_j(\vs)\}_{j = 1}^{n_u}$, such that $\phi_i(\vs_j) =
\delta_{ij}$. In this work these are exclusively the $C^0(\Omega)$ linear polynomial
``hat'' basis functions. This gives the finite-dimensional function space $V_h = \spans
\{\phi_j(\vs)\}_{j = 1}^{n_u}$, which is the space we look for solutions in.
Next, we rewrite $u$ and $v$ in terms of these basis functions:
$u_h(\vs, t) = \sum_{j = 1}^{n_u} u_j(t) \phi_j(\vs)$ and $v_h(\vs, t) = \sum_{j
  = 1}^{n_u} v_j(t) \phi_j(\vs)$. As the weak form must hold for all $v_h \in
V_h$, this is equivalent to holding for all $\phi_j$. Thus, the
weak form can now be rewritten in terms of this set of basis functions
\[
  \langle \partial_t u_h, \phi_j \rangle 
  + \cA_{\Lambda} (u_h, \phi_j) 
  + \langle F_\Lambda(u_h), \phi_j \rangle 
  = \langle f, \phi_j \rangle + \langle \dot{\xi}, \phi_j \rangle, 
  \quad j = 1, \ldots, n_u.
\]
Note that, in general, $u_h$ and $v_h$ do not necessarily have to be defined on the same function space, but as we use the linear basis functions in this work we stick with this here.

As stated in the main text, this is an \gls*{SDE} over the \gls*{FEM} coefficients
$\vu(t) = (u_1(t), \ldots, u_{n_u}(t))^\top$, given by
\[
  \mM \frac{\md \vu}{\md t} + \mA \vu + \cF(\vu) = \vb + \dot{\bm{\xi}},
  \quad \dot{\bm{\xi}}(t) \sim \NPDF(\bm{0}, \delta(t - t') \cdot \mG)
\]
where $\mM_{ij} = \langle \phi_i, \phi_j \rangle$, $\mA_{ij} = \cA_{\Lambda}
(\phi_i, \phi_j)$, $\cF(\vu)_j = \langle F_\Lambda(u_h), \phi_j \rangle$, $\vb_j
= \langle f, \phi_j \rangle$, and $\mG_{ij} = \langle \phi_i, \langle k(\cdot,
\cdot), \phi_j \rangle \rangle$. Letting $\mG = \mL \mL^\top$ we can then write
this in the familiar notation as above
\[
  \mM \md \vu + \mA \vu \md t + \cF(\vu) \md t = \vb \md t + \mL \, \md \mW(t),
\]
from which an Euler-Maruyama time discretisation gives
\[
  \vu_n = (\mI - \Delta_t \mM^{-1} \mA) \vu_{n - 1} - \Delta_t \mM^{-1}
  \cF(\vu_{n - 1}) + \Delta_t \mM^{-1} \vb + \mM^{-1} \mL \Delta \mW_{n - 1},
\]
where $\Delta \mW_{n - 1} \sim \mathcal{N}(0, \Delta_t \mI)$,
eventually defining a transition model of the form
\begin{align}
  p_{\Lambda}(\vu_n | \vu_{n-1})
  = \NPDF\left(
  (\mI - \Delta_t \mM^{-1} \mA) \vu_{n - 1}
  -\Delta_t \mM^{-1} \cF(\vu^{n - 1}) + \Delta_t \mM^{-1} \vb,
  \Delta_t \mM^{-1} \mG \mM^{-\top}
  \right).
\end{align}
Note that also that the \gls*{statFEM} methodology also allows for implicit
discretisations which may be desirable for time-integrator stability. The
transition equations for these approaches can be written out in closed form,
yet although they give Markovian transition models, the resultant transition densities
$p(\vu_n | \vu_{n - 1}, \mLambda)$ are not necessarily Gaussian due to the
nonlinear dynamics being applied to the current state $\vu_n$.
Letting $\ve_{n - 1} = \mL \Delta \mW_{n - 1} \sim \mathcal{N}(\mathbf{0}, \Delta_t \mG)$, then the implicit Euler is
\begin{equation}
  \label{eq:nl-implicit-euler}
  \mM \left( \vu_n - \vu_{n - 1} \right)
  + \Delta_t \mA \vu_{n}
  + \Delta_t \mathcal{F}(\vu_{n})
  + \Delta_t \vb = \ve_{n - 1},
\end{equation}
and the Crank-Nicolson is
\begin{equation}
  \label{eq:nl-crank-nicolson}
  \mM \left( \vu_n - \vu_{n - 1} \right)
  + \Delta_t \mA \vu_{n - 1/2}
  + \Delta_t \mathcal{F}(\vu_{n - 1/2})
  + \Delta_t \vb = \ve_{n - 1},
\end{equation}
where $\vu_{n - 1/2} = \left(\vu_n + \vu_{n - 1}\right)/2$. Furthermore, to
compute the marginal measure $p(\vu_n | \mLambda)$ this also requires integrating
over the previous solution $\vu_{n - 1}$; again due to nonlinear dynamics this
will not necessarily be Gaussian.

In each of these cases, therefore, the transition equation is
\[
  \mathcal{M}(\vu_n, \vu_{n - 1}) = \ve_{n - 1},
\]
where we take, for the implicit Euler,
\[
  \mathcal{M}(\vu_n, \vu_{n - 1}) =
  \mM \left( \vu_n - \vu_{n - 1} \right)
  + \Delta_t \mA \vu_{n}
  + \Delta_t \mathcal{F}(\vu_{n})
  + \Delta_t \vb
\]
and, for the Crank-Nicolson,
\[
  \mathcal{M}(\vu_n, \vu_{n - 1})=
  \mM \left( \vu_n - \vu_{n - 1} \right)
  + \Delta_t \mA \vu_{n - 1/2}
  + \Delta_t \mathcal{F}(\vu_{n - 1/2})
  + \Delta_t \vb.
\]
In practice due to conservative properties of the Crank-Nicolson discretisation
we use this for all our models in this work.

Discretised solutions $\vu_n$ are mapped at time $n$ to ``pseudo-observations''
via the observation process
\begin{align*}
  \vx_n = \mH \vu_n + \vr_n, \quad \vr_n \sim \NPDF(\bm{0},\mR).
\end{align*}
This observation process has the density $p_{\nu}(\vx_n | \vu_n)$ where $\nu =
\{\mH, \mR\}$. As stated in the main text, the pseudo-observation operator $\mH$
and observational covariance $\mR$ are assumed known in this work. We typically
use a diagonal covariance, setting $\mR = \sigma^2 \mI$. In the \gls*{PDE} case,
for a given \gls*{statFEM} discretisation as above, these pseudo-observations
are assumed to be taken on a user-specified grid, given by
$\vx_{\mathrm{obs}} \in \bR^{n_x}$. The pseudo-observation operator thus acts as
an interpolant, such that
\[
  \mH \vu_n = \left[u_h(\vx^1_{\mathrm{obs}}, n \Delta_t),
    u_h(\vx^2_{\mathrm{obs}}, n \Delta_t), \ldots, u_h(\vx^{n_x}_{\mathrm{obs}}, n \Delta_t)\right]^\top.
\]
For the \gls*{ODE} case, we have worked with observation operators that extract
relevant entries from the state vector.
The pseudo-observations are mapped to high-dimensional observed data through a
possibly nonlinear observation model which has the probability density
$p_\theta(\vy_n | \vx_n)$. Recall that in this, $\theta$ are neural network parameters. This
defines the decoding component of our model (see Figure~\ref{fig:network}).\\

\textbf{Nonlinear Filtering for Latent State Estimation.} To perform state
inference given a set of pseudo-observations we use the \acrfull*{ExKF}. The
\gls*{ExKF} constructs an approximate Gaussian posterior distribution via
linearising about the nonlinear model $\mathcal{M}(\cdot)$. The action
of the nonlinear $\mathcal{M}(\cdot)$ is approximated via tangent linear
approximation. We will derive our filter in the general
context of a nonlinear Gaussian \gls*{SSM} given by
\begin{align*}
  \text{Transition: }&\quad \mathcal{M}(\vu_n, \vu_{n-1}) = \ve_{n - 1}, \quad \ve_n \sim \mathcal{N}(\mathbf{0}, \mQ),\\
  \text{Observation: }&\quad \vx_n = \mH\vu_n + \vr_n, \quad \vr_n \sim \mathcal{N}(\mathbf{0}, \mR).
\end{align*}
This allows for the use of implicit time-integrators and subsumes the derivation
for the explicit case. We assume that at the previous timestep an
approximate Gaussian posterior has been obtained,
$p(\vu_{n-1} | \vx_{1:n-1}, \mLambda) = \mathcal{N}(\vm_{n-1}, \mC_{n-1})$. For each $n$
the \gls*{ExKF} thus proceeds as:
\begin{enumerate}
\item Prediction step. Solve $\mathcal{M}(\hat{\vm}_n, \vm_{n-1}) = 0$ for
  $\hat{\vm}_n$. Calculate the tangent linear covariance update:
  \[
    \hat{\mC}_n = \mJ_n^{-1} \left( \mJ_{n - 1} \mC_{n - 1} \mJ_{n - 1}^\top + \mQ \right) \mJ_n^{-\top},
  \]
  where $\mJ_n = \partial \mathcal{M} / \partial \vu_{n}|_{\hat{\vm}_n, \vm_{n-1}}$ and
  $\mJ_{n-1} = \partial \mathcal{M} / \partial \vu_{n-1}|_{\hat{\vm}_n, \vm_{n-1}}$.

  This gives $p(\vu_{n}|\vx_{1:n-1}, \mLambda) = \mathcal{N}(\hat{\vm}_{n}, \hat{\mC}_{n})$.
\item Update step. Compute the posterior mean $\vm_n$ and covariance $\mC_n$:
  \begin{align*}
    \vm_n &= \hat{\vm}_{n} + \hat{\mC}_{n}\mH^T (\mH \hat{\mC}_{n} \mH^T + \mR)^{-1} (\vy_n - \mH \hat{\vm}_{n}),\\
    \mC_n &= \hat{\mC}_{n} - \hat{\mC}_{n}\mH^T (\mH \hat{\mC}_{n} \mH^T + \mR)^{-1}\mH\hat{\mC}_{n}.
  \end{align*}
  This gives $p(\vu_{n}|\vx_{1:n}, \mLambda) = \NPDF(\vm_{n}, \mC_n)$.
\end{enumerate}

The log-marginal likelihood can be calculated recursively, with each term of the
log-likelihood computed after each prediction step:
\begin{align*}
  \log p(\vx_{1:N} | \mLambda) &= \log p(\vx_1| \mLambda) + \sum_{n=2}^{N} \log p(\vx_n|\vx_{1:n-1}, \mLambda), \\
  p(\vx_{n}|\vx_{1:n-1}, \mLambda) &= \mathcal{N}(\mH \hat{\vm}_n, \mH \hat{\mC}_{n} \mH^T + \mR).
\end{align*}
Note that although we focus on the \gls*{ExKF} other nonlinear filters could be
used; two popular alternatives are the ensemble Kalman filter~\citep{evensen2003ensemble},
or, the particle filter~\citep{doucet2000sequential}. For a linear dynamical model, such as the
advection equation considered in Section~\ref{sec:advection-results}, the
\gls*{ExKF} reduces to the standard Kalman filter~\citep{kalman1960new}.

\end{document}